\documentclass[10pt]{article}
\ifdefined\pdfminorversion\pdfminorversion=7\fi

\PassOptionsToPackage{table}{xcolor}
\usepackage[accepted]{tmlr}

\def\month{10}
\def\year{2026}
\def\openreview{\url{https://openreview.net/forum?id=TJyPRRrLIV}}

\usepackage{amsmath}
\input{t1lmr.fd}
\input{t1lmss.fd}
\DeclareFontShape{T1}{lmr}{bx}{sc}{<->ssub*lmr/bx/n}{}
\DeclareFontShape{T1}{lmss}{bx}{sc}{<->ssub*lmss/bx/n}{}
\usepackage[table]{xcolor}
\definecolor{Gray}{gray}{0.9}
\newcommand{\Ours}{\textsc{UnSAMv2}\xspace}
\newcommand{\ours}{\textsc{UnSAMv2}\xspace}
\newcommand{\Oursplus}{\textsc{UnSAMv2+}\xspace}

\def\tablefontsmall#1#{
    \fontsize{8}{10}
    #1
    \selectfont
}

\newcommand{\spplus}[1]{\scriptsize\bf\textcolor{Green}{#1}}

\usepackage{array}
\usepackage{subcaption}
\usepackage{multirow,multicol}
\usepackage[flushleft]{threeparttable}
\usepackage{comment}
\usepackage{algorithmic}
\usepackage{booktabs}
\usepackage{bbm, dsfont}
\usepackage[font=small,labelfont=bf]{caption}

\usepackage{amssymb}%
\usepackage{pifont}%
\usepackage{caption}

\newsavebox{\videobox}
\newsavebox{\vlmtablebox}

\newcommand{\cmark}{\text{\ding{51}}}%
\newcommand{\xmark}{\text{\ding{55}}}%

\newcommand{\tablestyle}[2]{\setlength{\tabcolsep}{#1}\renewcommand{\arraystretch}{#2}\centering\footnotesize}

\newcolumntype{x}[1]{>{\centering\arraybackslash}p{#1pt}}

\newlength\savewidth

\makeatletter\renewcommand\paragraph{\@startsection{paragraph}{4}{\z@}
  {.5em \@plus1ex \@minus.2ex}{-.5em}{\normalfont\normalsize\bfseries}}\makeatother

\usepackage{graphicx}
\usepackage{enumitem}
\usepackage{wrapfig}
\usepackage{lipsum}
\usepackage[table]{xcolor}
\usepackage{soul}
\usepackage{booktabs}
\usepackage{makecell}

\newcolumntype{H}{>{\setbox0=\hbox\bgroup}c<{\egroup}@{}}
\newcolumntype{a}{>{\columncolor{Gray}}c}

\usepackage{tabu}
\usepackage{xcolor}
\usepackage{nicematrix}

\definecolor{ForestGreen}{rgb}{0.13, 0.55, 0.13}
\definecolor{Green}{rgb}{0.0, 0.5, 0.0}
\definecolor{green(munsell)}{rgb}{0.0, 0.66, 0.47}
\definecolor{green(ryb)}{rgb}{0.4, 0.69, 0.2}
\definecolor{green(pigment)}{rgb}{0.0, 0.65, 0.31}
\definecolor{citecolor}{HTML}{0071bc}
\definecolor{GrayXMark}{gray}{0.7}
\definecolor{OracleTextColor}{gray}{0.55}

\definecolor{RevColor}{HTML}{000000}
\newcommand{\rev}[1]{{\color{RevColor}#1}}

\usepackage{needspace}

\usepackage{tabularx}
\usepackage[export]{adjustbox}

\usepackage{xspace}
\makeatletter
\DeclareRobustCommand\onedot{\futurelet\@let@token\@onedot}
\def\@onedot{\ifx\@let@token.\else.\null\fi\xspace}
\def\eg{\emph{e.g}\onedot}

\makeatother

\usepackage{hyperref}
\usepackage{url}
\usepackage[capitalize]{cleveref}

\title{\ours: Self-Supervised Learning Enables Segment \\ Anything at Any Granularity}

\author{\name Junwei Yu \email yujunwei04@berkeley.edu \\
      \addr UC Berkeley
      \AND
      \name Trevor Darrell \email trevordarrell@berkeley.edu \\
      \addr UC Berkeley
      \AND
      \name XuDong Wang \email xdwang@eecs.berkeley.edu \\
      \addr UC Berkeley, Duke University}

\begin{document}

{\hbadness=10000\maketitle}

\def\tabSemSAM#1{%
\begin{table}[#1]
\tablestyle{4pt}{1.05}
\footnotesize
\centering
\rev{%
\begin{tabular}{lrrrr}
Model (\# output masks) & PascalPart & PtIn-OOD & PACO & PtIn \\
\hline
Semantic-SAM~\citep{li2023semantic} (6) & 46.7 & 71.0 & 54.2 & 48.5 \\
\rowcolor{teal!5} UnSAMv2 (6 $g$)   & 48.4 & 72.5 & 55.1 & 51.2 \\
\rowcolor{teal!5} UnSAMv2+ (6 $g$)  & 50.6 & 73.8 & 59.2 & 54.9 \\
\rowcolor{teal!5} UnSAMv2+ (10 $g$) & \textbf{55.8} & \textbf{76.1} & \textbf{64.3} & \textbf{60.7} \\
\end{tabular}}
\caption{\rev{\textbf{Comparison with Semantic-SAM~\citep{li2023semantic}.}
We report per-target best IoU over each model's output set with identical clicks on PascalPart, PartImageNet-OOD / PartImageNet,
and PACO. \ours outperforms Semantic-SAM on all
evaluated benchmarks while learning granularity control from unlabeled data.}}
\label{tab:semsam-compare}
\end{table}%
}

\def\tabVideoQuantWrap{%
\Needspace*{8\baselineskip}%
\begin{wraptable}[4]{r}{0.42\linewidth}
\vspace{-12pt}
\captionsetup{font=footnotesize, skip=2pt}
\centering
\scriptsize
\setlength{\tabcolsep}{3pt}
\renewcommand{\arraystretch}{0.95}
\rev{%
\begin{tabular}{@{}l@{\hspace{5pt}}rrr@{}}
\hline
Method & J\&F & Anch.\ J\&F & TCE$\downarrow$ \\
\hline
SAM-2 & 57.6 & 47.0 & 11.15 \\
\rowcolor{teal!5} UnSAMv2 & \textbf{64.2} & \textbf{50.3} & \textbf{10.88} \\
\hline
\end{tabular}}
\caption{\rev{\textbf{Video results on DAVIS.}}}
\label{tab:video-quant}
\end{wraptable}%
}

\def\parImperfectGWrapBlock{%
\Needspace*{14\baselineskip}%
\rev{%
\begin{wraptable}[8]{r}{0.44\linewidth}
\vspace{-14pt}
\captionsetup{font=footnotesize, skip=2pt}
\centering
\scriptsize
\setlength{\tabcolsep}{3pt}
\renewcommand{\arraystretch}{0.95}
\begin{tabular}{@{}l@{\hspace{5pt}}r@{\hspace{7pt}}rrr@{}}
\hline
 & & \multicolumn{3}{c}{UnSAMv2} \\
\cline{3-5}
\rule{0pt}{2.3ex}Dataset & SAM-2 & $g^{*}$ & $\delta{=}0.05$ & $\delta{=}0.10$ \\
\hline
PascalPart   & 9.37 & {7.84} & {8.39} & {8.86} \\
PartImageNet & 5.09 & {4.21} & {4.57} & {4.92} \\
Berkeley     & 1.29 & {1.10} & {1.16} & {1.28} \\
GrabCut      & 1.48 & {1.16} & {1.28} & {1.34} \\
\hline
\end{tabular}
\caption{\textbf{Robustness to imperfect granularity.} NoC$_{80}$
($\downarrow$) with the granularity off the oracle $g^{*}$ by $\delta$.}
\label{tab:imperfect-g}
\end{wraptable}%
\noindent Beyond oracle-based results, Table~\ref{tab:imperfect-g} evaluates
the setting where the user's granularity input deviates from the oracle value.
With an off-oracle granularity input, \ours still selects the target object
with fewer clicks than SAM-2 on all benchmarks, since later point prompts
attend to the initial granularity embedding. 
\par}%
}

\def\parSemanticsWithTable{%
\Needspace*{14\baselineskip}%
\rev{%
\begin{wraptable}[6]{r}{0.33\linewidth}
\vspace{-12pt}
\captionsetup{font=footnotesize, skip=2pt}
\centering
\scriptsize
\setlength{\tabcolsep}{2.5pt}
\renewcommand{\arraystretch}{0.95}
\begin{tabular}{@{}l@{\hspace{4pt}}l@{\hspace{4pt}}r@{}}
\hline
VLM & Input & Spearman \\
\hline
Qwen3-VL-8B & Image + text & {+0.81} \\
Qwen3-VL-8B & Text only    & +0.75 \\
\hline
\end{tabular}
\caption{\textbf{Semantic alignment} between VLM-predicted and
oracle granularity (PartImageNet test).}
\label{tab:semantic-corr}
\end{wraptable}%
\noindent\textbf{Discussion on semantics of the granularity scale.}
Our granularity label is monotonic with mask size \emph{within a given
part--whole hierarchy} as shown in Eq.~\eqref{eq:gra_calc}, which normalizes each mask by the smallest and largest
masks of its own hierarchy, so two whole instances receive $g=1.0$ even if
their absolute areas differ substantially. This explains why the mean mask area
in Fig.~\ref{fig:gra-dist-wrap} rises with $g$ while its variance also grows --- the same
whole-instance granularity may correspond to a small person or a large boat.
To probe semantic alignment directly, Table~\ref{tab:semantic-corr} measures
the Spearman correlation between a frozen VLM's granularity prediction from the
ground-truth phrase (e.g., ``head of the fish'') and the oracle granularity of
\ours; the strong correlation ($+0.81$ with image and text, $+0.75$ text-only)
shows the learned scale aligns with part--whole semantics rather than absolute
area.\par}%
}

\def\tabRepPseudoSubtables#1{%
\begin{table}[#1]
\begin{minipage}[t]{0.46\textwidth}\vspace{0pt}
\tablestyle{3pt}{1.05}
\footnotesize
\centering
\rev{%
\begin{tabular}{l@{\hspace{5pt}}rrr}
Granularity repr. & NoC$_{80}\downarrow$ & NoC$_{85}\downarrow$ & 1-IoU \\
\hline
10 discrete bins        & 6.89 & 8.36 & 49.2 \\
Continuous, MLP proj.   & 6.16 & 7.47 & 53.6 \\
\rowcolor{teal!5}
\textbf{Cont., Fourier (ours)} & \textbf{6.03} & \textbf{7.44} & \textbf{54.3} \\
\end{tabular}}
\captionof{table}{\rev{\textbf{Ablations on granularity representation.}
We report mean results on PascalPart and PartImageNet across different design choices.}}
\label{tab:gra-rep-ablation}
\end{minipage}
\hfill
\begin{minipage}[t]{0.50\textwidth}\vspace{0pt}
\tablestyle{3pt}{1.05}
\footnotesize
\centering
\rev{%
\begin{tabular}{l@{\hspace{5pt}}rrr}
Pseudo-label design & NoC$_{80}\downarrow$ & NoC$_{85}\downarrow$ & 1-IoU \\
\hline
Absolute area, no hierarchy & 7.61 & 8.99 & 44.3 \\
Hierarchical, no sqrt       & 6.69 & 8.12 & 49.1 \\
Divide only, no conquer     & 8.01 & 9.45 & 42.7 \\
\rowcolor{teal!5}
\textbf{Hier.\ + sqrt + d\&c (ours)} & \textbf{6.03} & \textbf{7.44} & \textbf{54.3} \\
\end{tabular}}
\captionof{table}{\rev{\textbf{Ablations on pseudo-label design.}
We report average results over PascalPart and PartImageNet.}}
\label{tab:pseudo-label-ablation}
\end{minipage}
\end{table}%
}

\def\tabWholeImageCounts#1{%
\begin{table}[#1]
\tablestyle{4pt}{1.1}
\footnotesize
\centering
\rev{%
\begin{tabular}{l@{\hspace{8pt}}rr}
Model & Masks generated & w/ conf.\ filtering + NMS \\
\hline
SAM-2 (3 mask tokens per grid point)       & 635   & -- \\
UnSAMv2 (10 $g$ values per grid point)     & 1,206 & 714 \\
\end{tabular}}
\caption{\rev{\textbf{Whole-image proposal counts per image.}
We prompt \ours with a $64{\times}64$ point grid at each
$g \in \{0.1, 0.2, \ldots, 1.0\}$, keep masks with confidence score $>0.5$, and
deduplicate with NMS at IoU threshold $0.7$. Filtering and deduplication reduce
the pool from over 1{,}200 to about 700 masks, enabling a fair recall comparison
with the baseline.}}
\label{tab:whole-image-counts}
\end{table}%
}

\def\tabNoveltyCompare#1{%
\begin{table}[#1]
\tablestyle{4pt}{1.1}
\footnotesize
\centering
\rev{%
\begin{tabular}{l@{\hspace{8pt}}ccccc}
Property & UnSAM & SOHES & GraCo & SAM & \textbf{UnSAMv2} \\
\hline
Unsupervised            & \cmark & \cmark & \xmark & \xmark & \cmark \\
Granularity input       & \xmark & \xmark & \cmark & \xmark & \cmark \\
Continuous granularity  & \xmark & \xmark & \xmark & \xmark & \cmark \\
Hierarchical perspective& \cmark & \xmark & \xmark & \xmark & \cmark \\
\end{tabular}}
\caption{\rev{\textbf{Technical comparison with prior work}
(UnSAM~\citep{wang2024segment}, SOHES~\citep{cao2024sohes},
GraCo~\citep{zhao2024graco}, SAM~\citep{kirillov2023segment}).
\ours is the first to learn a \emph{continuous} granularity variable from
unsupervised part--whole hierarchies.}}
\label{tab:novelty-compare}
\end{table}%
}

\def\secRebuttalAppendix{%
\rev{\section{Additional discussions}\label{sec:rebuttal-appendix}}

\rev{\subsection{Technical Comparison with Prior Work}
Table~\ref{tab:novelty-compare} summarizes the technical differences from prior
granularity-aware and unsupervised segmentation methods. To further demonstrate the effectiveness of our granularity-controllable \ours, we compare with multi-granularity baseline Semantic-SAM in Table~\ref{tab:semsam-compare}. With
identical clicks and per-target best IoU over each model's output set, \ours
outperforms it on all tested benchmarks.}
\tabNoveltyCompare{ht}
\tabSemSAM{ht}

\rev{\subsection{Biases and Failure Modes of the Pseudo-Label Pipeline}
Our pseudo-label pipeline described in Sec~\ref{sec:method-dico} carries a grouping bias inherited from the
self-supervised features and the MaskCut algorithm. We observe two recurring
artifacts: MaskCut-based discovery does not always produce smooth object
boundaries, which can introduce small spurious masks during merging, and
strongly textured surfaces tend to be over-segmented into several fragments.
We mitigate both by tuning the merging thresholds $\theta$ in the conquer stage
and raising the confidence threshold $\tau$ in the divide stage so the initial
divide-stage masks are as clean as possible.}

\rev{\subsection{Whole-Image Segmentation Evaluation Protocol}
For whole-image segmentation results, we use AR-1000 as our primary metrics. We present the details on how we derive \ours masks on evaluation sets. For each image we prompt \ours with a $64{\times}64$ point grid at each of the
ten granularity values $g\in\{0.1,\dots,1.0\}$, keep masks with confidence
score $>0.5$, and deduplicate the union with NMS at IoU $0.7$
(Table~\ref{tab:whole-image-counts}). The resulting proposal count (714) is
comparable to the SAM-2 baseline (635), enabling a fair AR comparison.}
\tabWholeImageCounts{ht}
}

\vspace{2mm}

\begin{abstract}
The Segment Anything Model (SAM) family has become a widely adopted vision foundation model, but its ability to control segmentation granularity remains limited. 
Users often need to refine results manually — by adding more prompts or selecting from pre-generated masks — to achieve the desired level of detail. This process can be ambiguous, as the same prompt may correspond to several plausible masks, and collecting dense annotations across all granularities is prohibitively expensive, making supervised solutions infeasible. 
To address this limitation, we introduce \ours which enables segment anything at any granularity without using any human annotations. 
\Ours extends the divide-and-conquer strategy of UnSAM by discovering abundant mask-granularity pairs and introducing a novel granularity control embedding that enables precise, continuous control over segmentation scale. 
Remarkably, with only 6$K$ unlabeled images and 0.02\% additional parameters, \Ours substantially enhances SAM-2, achieving segment anything at any granularity across interactive, whole-image, and video segmentation tasks. Furthermore, we introduce a language-guided segmentation pipeline that treats granularity as a {predictable} variable by grounding it in text semantics via an external vision-language model, improving previous SOTA in 1-IoU (40.4 $\rightarrow$ 47.3). Across 11+ benchmarks, \Ours achieves consistent gains over SAM/SAM-2 baselines, improving $\text{NoC}_{90}$ (5.69 $\rightarrow$ 4.75), 1-IoU (58.0 $\rightarrow$ 73.1), and $\text{AR}_{1000}$ (49.6 $\rightarrow$ 68.3) by a large margin. Our work demonstrates that a small amount of unlabeled data can deliver large, practical improvements to strong supervised baselines! 
Our code, dataset, and demo are available at \url{https://github.com/yujunwei04/UnSAMv2}.

\end{abstract}

\def\figTeaser#1{
    \begin{figure}[#1]
      \centering
      \includegraphics[width=0.99\linewidth]{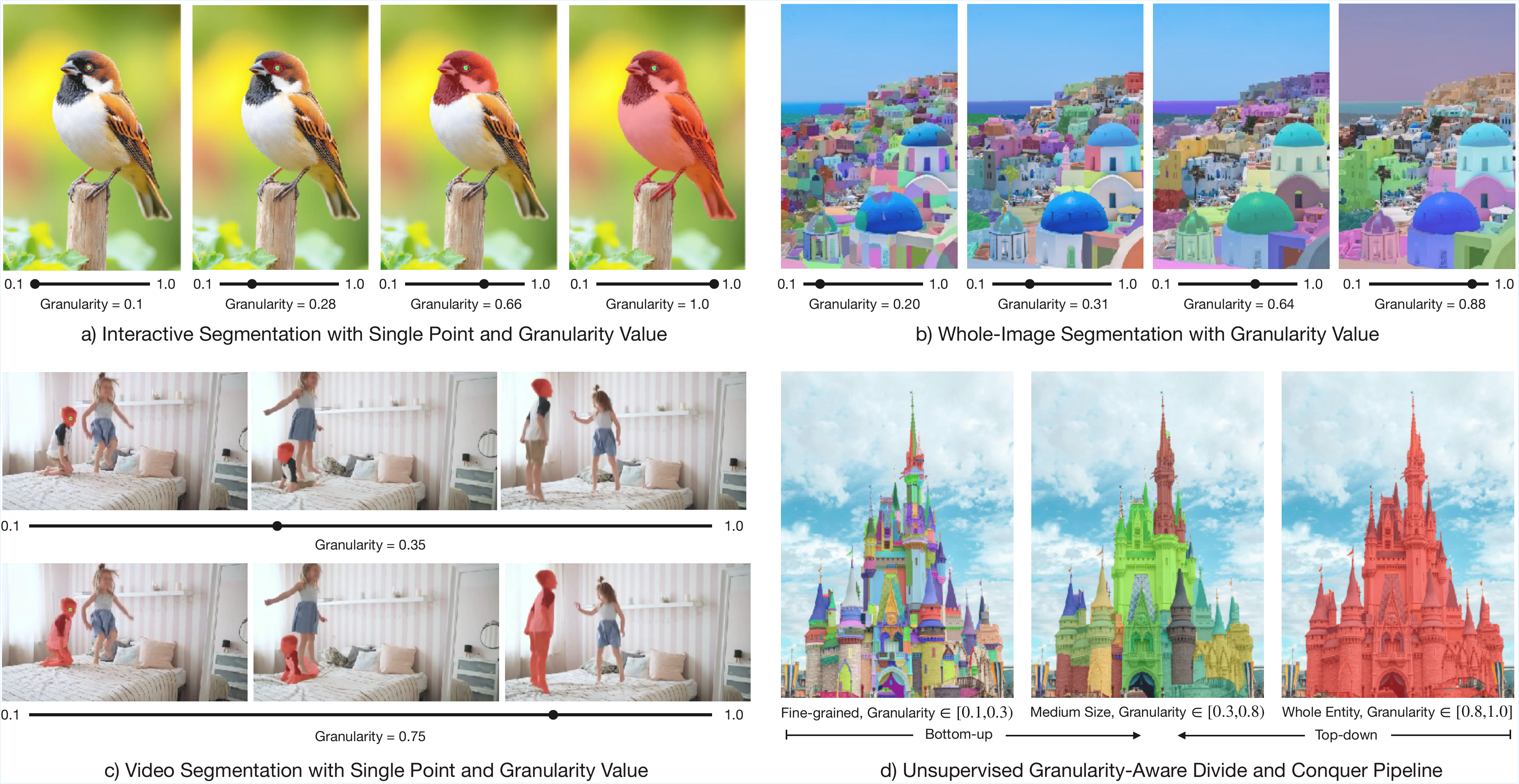}\vspace{-3pt}
    \caption{
        What is an object? The notion has long been debated: should it follow a model's learned semantics or an annotator's subjective judgment?
        \Ours takes a third path, granting users full flexibility to define objectness through promptable segmentation with a single point and a continuous, controllable granularity score.
        Built on SAM-2 \citep{ravi2024sam}, \Ours introduces a granularity-aware, self-supervised training pipeline based on divide-and-conquer pseudo-labels \citep{wang2024segment}. 
        Trained on just 6000 unlabeled images, it segments anything from fine-grained parts to holistic objects, achieving state-of-the-art performance across interactive, whole-image, and video segmentation tasks.
    } \label{fig:gra-dist}
    \end{figure}
}

\def\figSAMGraDistAB#1{
  \begin{figure}[#1]
    \captionsetup{font=small}
    \centering

    \begin{minipage}[t]{0.58\linewidth}
      \centering
      \includegraphics[width=\linewidth]{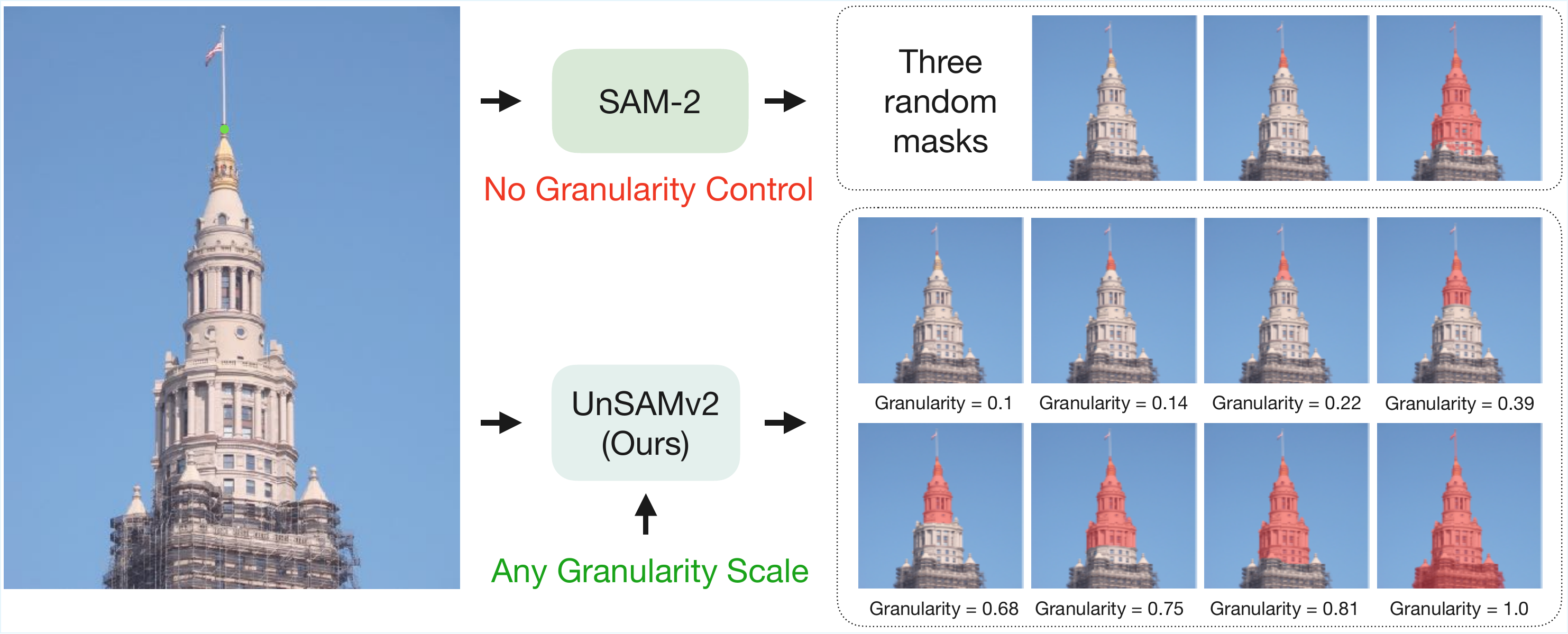}\vspace{-2pt}\\
      {\bf (a)} From ambiguity to control
    \end{minipage}
    \hfill
    \begin{minipage}[t]{0.39\linewidth}
      \centering
      \includegraphics[width=\linewidth]{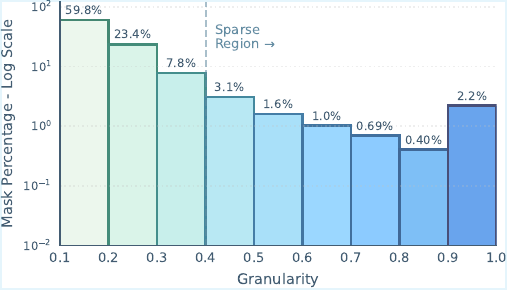}\vspace{-2pt}\\
      {\bf (b)} Granularity distribution
    \end{minipage}

    \vspace{-2pt}
    \caption{
      \textbf{From ambiguity to control, and learning across scales.}
      \textbf{(a)} Without granularity input, SAM-2 may output multiple plausible masks per point; \ours introduces a continuous granularity variable to retrieve the intended object at any scale with a single prompt.
      \textbf{(b)} Our unsupervised divide-and-conquer pipeline yields a rich, left-tailed granularity distribution dominated by fine-grained pseudo-masks, yet \Ours learns stable semantics across all scales.
    }
    \label{fig:sam-gra-dist}
  \end{figure}
}

\section{Introduction}
\label{sec:intro}

\figTeaser{t!}

What is an object?  
This question has long been debated in both vision and cognition. Should an object be defined by a model’s learned semantics or by an annotator’s subjective judgment?  
We propose a third perspective: granting users full flexibility to define objectness through a single point prompt and a continuous granularity score.

Imagine clicking on a single point in an image and smoothly adjusting a slider that controls segmentation granularity.  
At low granularity values, the model reveals fine-grained parts with precise boundaries; as the granularity increases, it gradually merges regions into larger and more semantically coherent entities. 
In essence, this transforms SAM’s~\citep{kirillov2023segment} three discrete mask hypotheses into a continuous granularity axis capable of producing multiple masks per point, each corresponding to a user-defined granularity. 
Such controllable segmentation enables users to flexibly define what constitutes an ``object'' for their usage, whether it involves part-level analysis, instance grouping, or large-scale region editing.

The emergence of the Segment Anything family~\citep{kirillov2023segment, ravi2024sam} has positioned SAM as a vision foundation model, significantly advancing diverse tasks such as video object tracking~\citep{yang2023track, ye2025entitysam}, multi-scale perception~\citep{huang2025segment, wang2025mtsam}, and compositional understanding~\citep{doersch2024bootstap, wang2023seggpt, chen2025conformalsam}.  
However, SAM and its successors rely heavily on \textit{supervised learning} from SA-1B dataset~\citep{kirillov2023segment}. 
This pipeline directly ties the model’s notion of an ``object'' to human annotation bias and limits its output to three discrete mask hypotheses per prompt.  
As a result, segmentation in SAM is \textit{defined by supervision} rather than \textit{discovered from data}.  
This design assumes a shallow hierarchy of objectness, while real-world scenes exhibit complex, nested part–whole structures that cannot be captured through fixed human labels.  
Although SA-1B covers a broad range of object sizes, it lacks explicit correspondences between instance- and part-level masks, making it difficult for supervised models to learn how granularity should vary continuously.

\figSAMGraDistAB{t}

\textit{We argue that segmentation granularity should be learned through \textit{unsupervised learning}, which allows models to infer object hierarchies directly from image statistics instead of depending on predefined labels.} 
To this end, we introduce \textbf{\ours}, a self-supervised framework that enables segmentation at any granularity without human supervised labels.
\Ours\ learns a continuous representation of granularity that bridges the gap between parts and wholes, giving users full control over segmentation masks.

\Ours\ builds upon UnSAM~\citep{wang2024segment}, which introduced an unsupervised divide-and-conquer strategy for discovering hierarchical masks.  
While UnSAM focused on unsupervised hierarchy construction, \textbf{\ours} extends this idea to \textit{granularity-controllable segmentation}.  
In the \emph{divide} stage, we use the normalized-cut method MaskCut~\citep{wang2023cut} to extract instance-level masks.
In the \emph{conquer} stage, we recursively merge similar pixels within each instance to discover finer parts, forming hierarchical pseudo-labels that encode relative scale.  
From these hierarchies, we compute a continuous granularity scalar for each mask, representing its position along the part--whole continuum.

We then augment SAM-2~\citep{ravi2024sam} with a lightweight granularity encoder and a granularity-aware mask token. 
Given a point prompt and a granularity scalar $g$, \Ours\ predicts the mask corresponding to the desired granularity, turning segmentation into a controllable function of scale.  
Training only the lightweight SAM-2 decoder for four hours with 2 A100 GPUs on 6,000 unlabeled images (0.02\% additional parameters) enables smooth interpolation between coarse and fine structures and reveals the latent hierarchy within SAM’s feature space.  

Across interactive, whole-image, and video segmentation benchmarks, \Ours\ consistently surpasses SAM-2~\citep{ravi2024sam} and previous state-of-the-art promptable segmentation methods.  
Evaluated on more than 11 widely used datasets, including SA-1B, COCO, and PartImageNet, \Ours\ improves $\text{NoC}_{90}$ from 5.69 to 4.75, 1-IoU from 58.0 to 73.1, and AR$_{1000}$ from 49.6 to 68.3, all achieved using only unlabeled data.  
The resulting model empowers users to define their own notion of objectness and to explore segmentation as a continuous, controllable process rather than a static prediction. We further demonstrate the practical utility of granularity control when combined with semantic prompts. Inspired by previous language-guided segmentation methods such as LISA~\citep{lai2024lisa}, SA2VA~\citep{yuan2025sa2va}, and SAM-3~\citep{carion2025sam}, we pair our granularity control with an external VLM that predicts an appropriate granularity value from the user’s text prompt. This improves the strongest baseline~\citep{yuan2025sa2va} from 40.41 to 47.29 in 1-IoU, demonstrating that granularity can be effectively grounded in text semantics.

\section{Related Work}
\label{sec:related}

\noindent \textbf{Multi-Granularity Segmentation.}
SAM~\citep{kirillov2023segment, ravi2024sam, carion2025sam} has greatly advanced segmentation performance by leveraging large-scale human-annotated data and extensive compute. Semantic-SAM~\citep{li2024segment} improves fine-grained predictions through a multiple-choice design~\citep{guzman2012multiple, Li_2018_CVPR}. However, these approaches constrain point-prompt predictions to a fixed number of candidate masks, forcing users to manually select from limited outputs or give additional prompts. This restriction highlights the need for explicit control over mask granularity. Recent work~\citep{wang2024order, lin2025refcut, li2023multi} has begun to tackle such ambiguity. GARField~\citep{kim2024garfield} and SAMPart3D~\citep{yang2024sampart3d, yang2025omnipart} address scale ambiguity in 3D scene decomposition via absolute scale conditioning, while GraCo~\citep{zhao2024graco} achieves granularity-controllable interactive segmentation by extending SimpleClick~\citep{liu2023simpleclick} with discrete granularity inputs. In contrast, \ours tackles mask ambiguity in a fully self-supervised manner by treating granularity as a continuous, relative concept within a hierarchical structure.

\noindent \textbf{Self-Supervised Learning and Unsupervised Segmentation.}
SSL methods such as MAE~\citep{he2022masked}, JEPA~\citep{assran2023self}, and DINO~\citep{caron2021emerging, oquab2023dinov2, simeoni2025dinov3} demonstrate that large-scale pretraining can endow vision transformers with strong representations, benefiting a wide range of downstream tasks~\citep{jiang2025rayzer, gao2024partgs, zhang2025concerto, depth_anything_v1, depth_anything_v2, wu2025sonata, karmann2025repurposing}. In parallel, unsupervised segmentation has gained lots of attention~\citep{wang2023tokencut, li2023paintseg, wang2022freesolo, simeoni2021localizing, kim2024eagle, hamilton2022unsupervised, arica2024cuvler, Niu_2024_CVPR, zhou2020matnet, hahn2025scene, karmann2025repurposing, sick2025cuts3d}. 
CutLER~\citep{wang2023cut,wang2024videocutler} greatly advanced unsupervised instance segmentation by using MaskCut~\citep{shi2000normalized} to iteratively extract objects from images. SOHES~\citep{cao2024sohes} adopts a bottom-up merging scheme, grouping pixels based on cosine similarity to progressively discover objects. Recently, UnSAM~\citep{wang2024segment} introduced a divide-and-conquer paradigm to generate hierarchical pseudo labels.  Building on these efforts, UnSAMv2 leverages the hierarchical mask discovery perspective and extends it to assign an explicit granularity scale for each pseudo mask. This enables unsupervised segmentation at any granularity level of detail explicitly.

\noindent \textbf{Language-Guided Segmentation.}
Language-guided segmentation~\citep{xia2024gsva,ren2024pixellm,wei2025instructseg,chen2024sam4mllm,bai2024one,wang2024segllm} grounds text semantics into pixel masks by combining VLMs with segmenters.
LISA~\citep{lai2024lisa} decodes a segmentation token from an MLLM into masks, while Sa2VA~\citep{yuan2025sa2va} and SAM-3~\citep{carion2025sam} further integrate SAM-style models with language for image/video prompting. We are the first to treat \emph{granularity} as a predictable variable by introducing a VLM-based pipeline that infers granularity from text semantics and combines it with granularity-controllable segmentation.

\def\figPipe#1{
    \begin{figure*}[#1]
      \centering
      \includegraphics[width=0.97\linewidth]{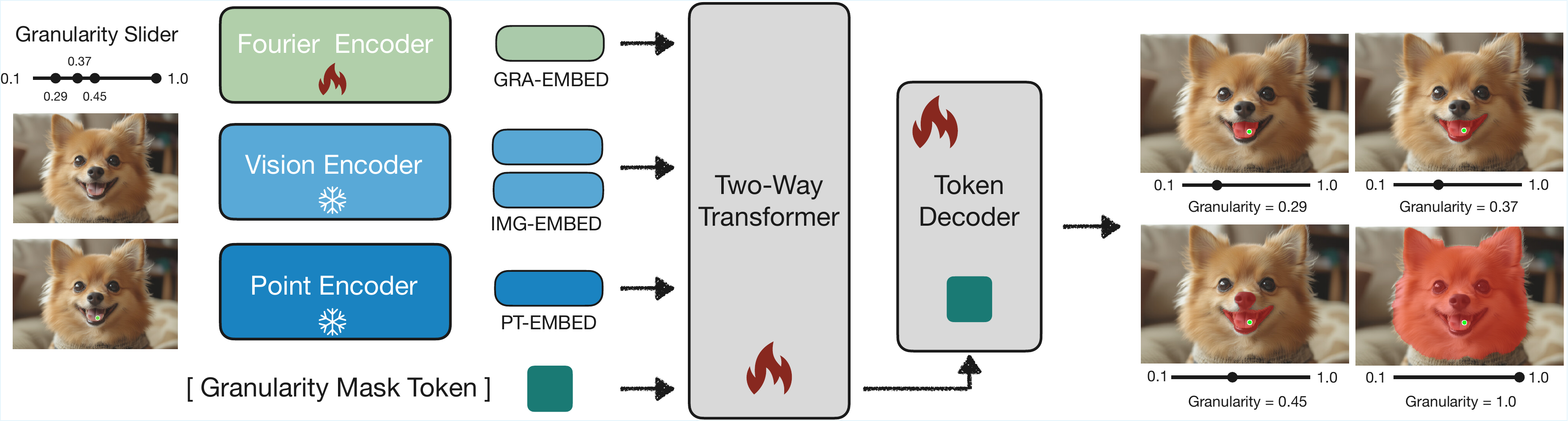}
      \vspace{-3pt}
        \caption{\textbf{Architecture of \ours.} Built on SAM-2, \Ours introduces a Fourier-based granularity encoder and a mask token to enable segmentation at any granularity. A scalar granularity input $g\!\in\![0.1,1]$ is mapped to a high-dimensional embedding via Fourier transformation and an MLP, then injected into the transformer alongside the sparse point embedding and dense image embedding. The granularity-aware mask token attends to image, point, and granularity embeddings, and is finally decoded by a token decoder into a mask at the requested granularity.}
      \label{fig:pipe}
    \end{figure*}
}

\def\figVLMPipe#1{%
  \begin{figure*}[#1]
    \captionsetup{font=small}
    \centering

    \begin{minipage}[t]{0.60\linewidth}\vspace{0pt}
      \centering
      \includegraphics[width=\linewidth]{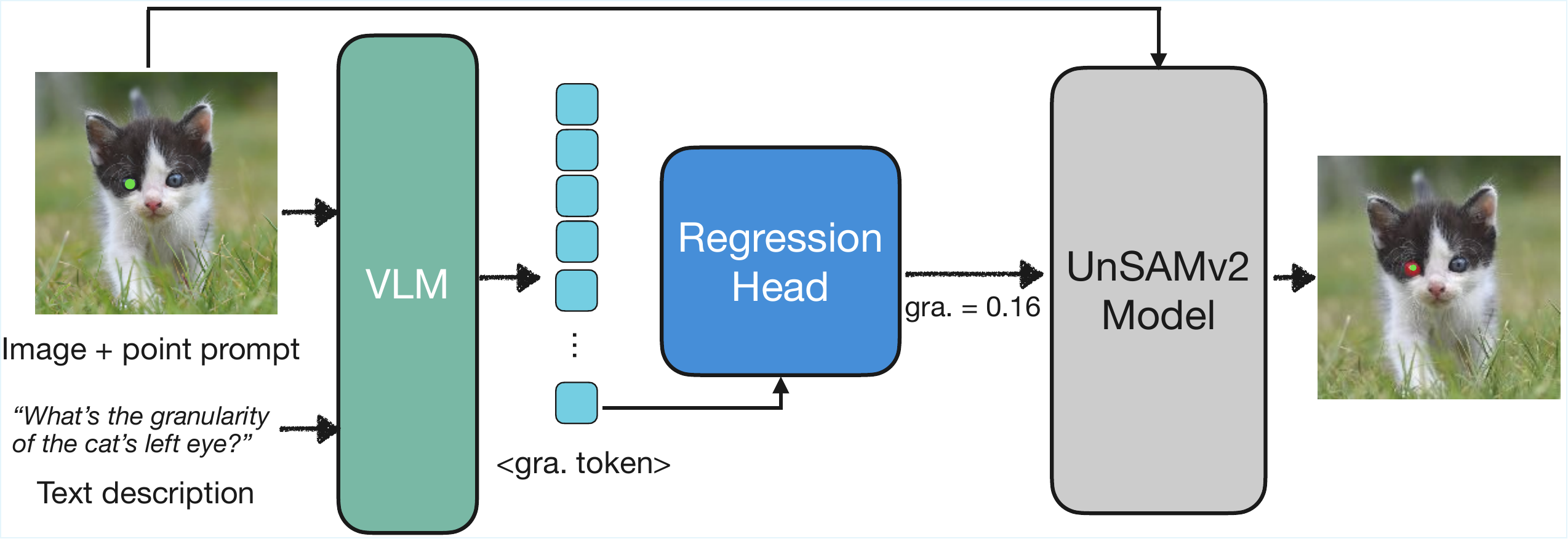}
    \end{minipage}
    \hfill
    \begin{minipage}[t]{0.36\linewidth}\vspace{0pt}
      \captionsetup{width=\linewidth, justification=raggedright, singlelinecheck=false}
      \captionof{figure}{\textbf{Language-guided segmentation.}
      Conditioned on the image and the text description, the VLM infers a granularity value, which is then provided to \Ours for segmentation.}
      \label{fig:vlm_pipe}
    \end{minipage}

  \end{figure*}%
}

\def\figGracoOurs#1{
    \begin{figure*}[#1]
    \centering
    \includegraphics[width=0.99\linewidth]{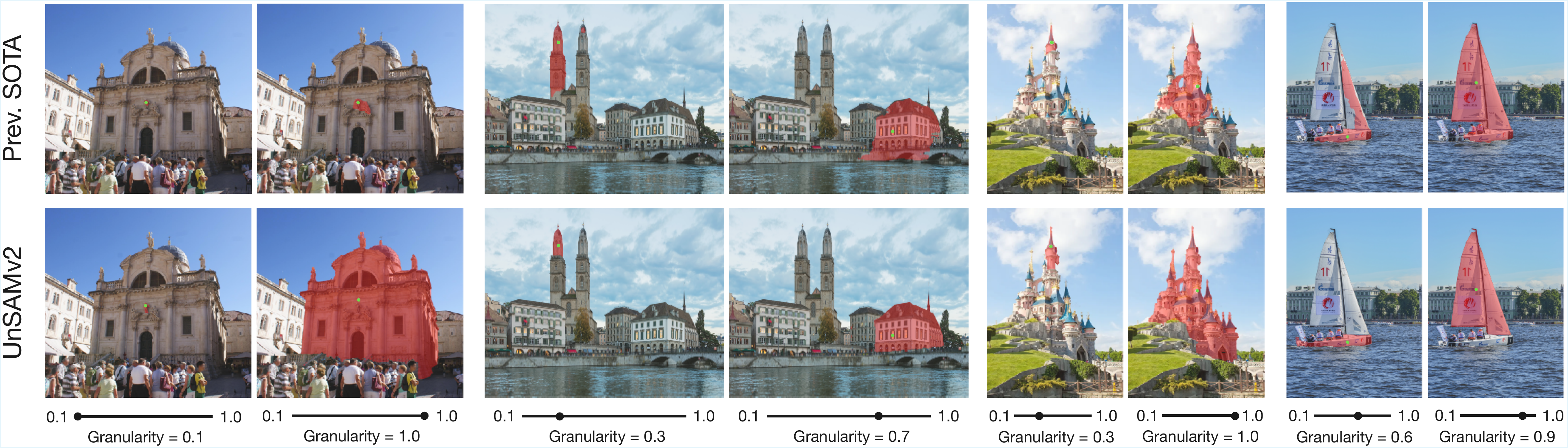}\vspace{-3pt}
    \caption{
    \textbf{Qualitative comparison with the previous state-of-the-art method}~\citep{zhao2024graco}. 
    Each scene shows results at different target granularity values. 
    Prior methods often break one object into parts at high granularity or include extra regions at low granularity. In contrast, \ours produces clear and consistent masks with smooth transitions across scales.
    }
    \label{fig:graco-ours}
    \end{figure*}
}

\def\figOursWholeSeg#1{
    \begin{figure*}[#1]
      \centering
      \includegraphics[width=0.99\linewidth]{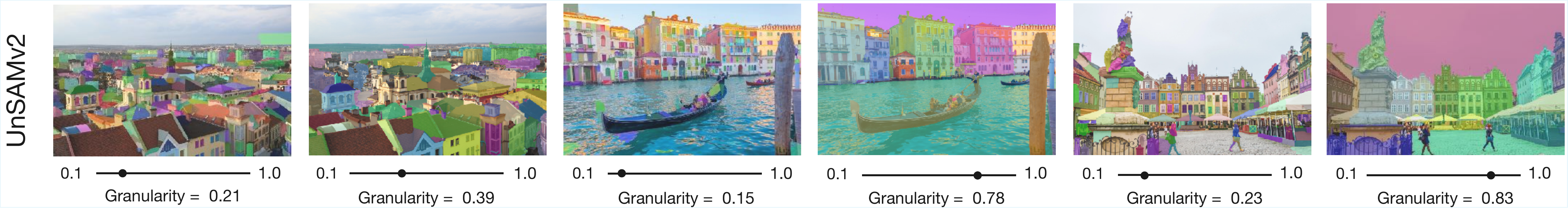}\vspace{-5pt}
\caption{
\textbf{Visualizations for whole-image segmentation.}
Low granularity reveals fine parts, while higher values recover whole objects. \ours offers controllable whole-image segmentation, even for scenes with many densely packed entities.}
    \label{fig:ours-whole-seg}
    \end{figure*}
}

\def\figGraOneAnaRela#1{
    \begin{figure}[#1]
    \centering
    \includegraphics[width=0.99\linewidth]{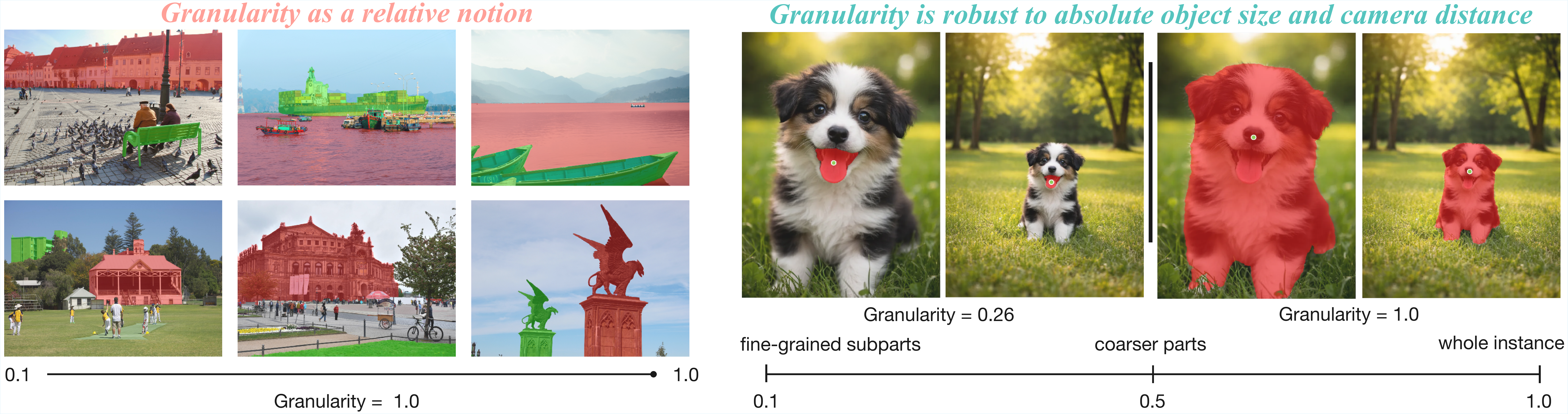}\vspace{-3pt}
    \caption{\textbf{Granularity is a relative notion and robust to camera distance.} At a fixed granularity value, mask sizes vary widely across scenes, showing that \ours learns granularity relationally, consistent with human perception of parts and wholes rather than simply associating it with absolute size.}
    \label{fig:gra-1-analysis}
    \end{figure}
}

\section{Method}

We present \textbf{\Ours}, a self-supervised framework that enables segmentation at arbitrary levels of granularity without human annotations.  
Unlike the supervised SAM~\citep{kirillov2023segment} pipeline, which depends on human-labeled object masks, \Ours\ learns granularity directly from image statistics through a hierarchy-aware divide-and-conquer process.  
This enables segmentation granularity to be continuously controlled by a single scalar input, rather than restricted to a fixed number of discrete mask tokens.  

We first review prior unsupervised segmentation methods (\cref{sec:method-prelim}) and the limitations of supervised training paradigms (\cref{sec:method-limitations}).  
We then present our granularity-aware divide-and-conquer algorithm that automatically constructs mask–granularity pairs from unlabeled data (\cref{sec:method-dico}).  
Next, we describe the architectural design that empowers any promptable segmentation model to interpret and control segmentation granularity (\cref{sec:method-model}).  
We then briefly discuss the difference with prior supervised learning works (\cref{sec:method-distinctions}).  
Then, we present \Oursplus, a lightly supervised variant that integrates SA-1B annotations to further refine the granularity learning process (\cref{sec:method-unsam+}). Finally, we associate  granularity with semantics using vision-language models (\cref{sec:gra-vlm}).

\subsection{Preliminaries}
\label{sec:method-prelim}

\noindent\textbf{UnSAM, MaskCut, and SOHES.}
UnSAM~\citep{wang2024segment} introduced a divide-and-conquer strategy to generate pseudo masks without supervision.  
In the \emph{divide} stage, a cut-based segmentation method MaskCut~\citep{wang2023cut} is applied to obtain instance-level masks.  
Then, inspired by bottom-up hierarchical segmentation, \eg, SOHES~\citep{cao2024sohes}, the \emph{conquer} stage iteratively merges similar pixels under a sequence of thresholds, constructing part–whole hierarchies.  
Formally, for a local image region $I_{\text{local}}$, patch-level features
$
    K = \text{DINO}(I_{\text{local}})
$
are extracted using DINO~\citep{simeoni2025dinov3}.  
Neighboring patches are merged according to the cosine similarity of their DINO features against thresholds $\theta_1, \ldots, \theta_l$, yielding part masks nested inside instances.  
This process produces a rich hierarchy of mask granularity.

\noindent\textbf{Segment Anything family.}
Segment Anything models~\citep{kirillov2023segment, ravi2024sam,carion2025sam} have advanced promptable segmentation with a scalable encoder–decoder design.  
\textbf{(1) Image encoder:} a ViT that maps an input image to multi-scale dense embeddings while preserving spatial structure.  
\textbf{(2) Prompt encoder:} embeddings for user inputs such as points, boxes, or masks that condition the segmentation process and guide attention to regions of interest.  
\textbf{(3) Mask decoder:} a lightweight transformer that fuses image and prompt features to predict segmentation masks.

Despite these strengths, the training pipeline is fully \emph{supervised} on SA-1B~\citep{kirillov2023segment}, which ties the notion of objectness to human-labeled masks.  
Moreover, the decoder employs three fixed mask tokens (small, medium, large), producing at most three hypotheses per prompt.  
This discretization limits controllable granularity and discourages hierarchical reasoning about parts and wholes, motivating an unsupervised formulation that learns granularity from data rather than from fixed labels.

\subsection{Limitations of SAM’s Supervised Paradigm}
\label{sec:method-limitations}

\paragraph{Lack of granularity control.}
When a single point corresponds to multiple plausible objects (\eg, a part versus the whole), SAM generates up to three discrete masks and requires manual selection by the user.  
Without an explicit granularity variable, the model cannot traverse scales continuously—fine details and coarse structures remain disconnected.  
This limitation not only reduces efficiency in interactive segmentation but also prevents smooth, interpretable control over the level of detail.

\paragraph{Lack of hierarchical reasoning.}
Supervised training on human-labeled masks encourages SAM to learn a flat object representation, where parts and instance-level segments are treated as isolated entities rather than components within a hierarchy.  
As a consequence, SAM lacks structural awareness and fails to capture relationships across scales.  
It struggles to segment scenes at intermediate levels of detail or to uncover the nested hierarchical structure of visual scenes (\cref{fig:sam-gra-dist}a).  
This limitation underscores the necessity of \textit{unsupervised learning}, which can recover hierarchical dependencies directly from image statistics rather than relying on human annotations.

\subsection{Granularity-Aware Divide-and-Conquer}
\label{sec:method-dico}
\figGraOneAnaRela{t!}

We extend UnSAM’s divide-and-conquer framework to automatically construct dense mask–granularity pairs from unlabeled images.  
The process consists of four stages.

\paragraph{Stage 1: Instance discovery via N-Cuts.}  
We employ the normalized-cut-based method CutLER~\citep{wang2023cut} to generate segmentation masks $\mathcal{M} = \{ m_1, \dots, m_n\}$ at varying levels of granularity. We filter out noisy CutLER mask outputs with a confidence threshold $\tau_{\text{conf}}$,
\begin{align}
\mathcal{M}_{\text{high}} = \{ m_i \mid \text{conf}(m_i) \geq \tau_{\text{conf}} \}, \forall i \in [n].
\end{align}

\paragraph{Stage 2: Instance–part relationship discovery.}  
We identify instance-level masks $\mathcal{M}_{\text{inst}} \subseteq \mathcal{M}_{\text{high}}$ based on two criteria:  
(i) their area-to-image-area ratio exceeds the threshold $\tau_{\text{area}}$, and
(ii) they dominate overlapping masks, which means for any $m_i \in \mathcal{M}_{\text{inst}}$ and $m_j \in \mathcal{M}_{\text{high}}$, we have
\begin{align}
\text{IoU}(m_i, m_j) \ge \tau_{\text{overlap}} \Rightarrow \text{Area}(m_i) \ge \text{Area}(m_j).
\end{align}
The remaining masks form $\mathcal{M}_{\text{rest}} = \mathcal{M}_{\text{high}} \setminus \mathcal{M}_{\text{inst}}$.  
Each $m_r \in \mathcal{M}_{\text{rest}}$ is assigned as a part of $m_i$ if $\text{IoU}(m_r, m_i) > \tau_{\text{overlap}}$; otherwise, it is discarded.  
Therefore, for every instance mask $m_i$, we have a set of part-level masks $\mathcal{M}_{i,\text{part}}$.

\paragraph{Stage 3: Fine-grained mask discovery.}  
To enrich granularity, we further apply patch merging to each $m_i \in \mathcal{M}_{\text{inst}}$, decomposing it into finer structures.  
The resulting fine-grained masks $\mathcal{M}_{i,\text{conquer}}$ are merged with $\mathcal{M}_{i,\text{part}}$ through Non-Maximum Suppression (NMS), denoted as $\mathcal{M}_{i,\text{final}}$.
This process expands granularity richness in a hierarchical perspective while maintaining mask quality.

\paragraph{Stage 4: Continuous granularity assignment.}  
After obtaining the instance and part hierarchies, we assign each mask $m_i \in \mathcal{M}_{i,\text{final}}$ a continuous granularity score $g_i \in [0.1,1.0]$ according to its relative area within the hierarchy:
\begin{align}
g_i = \left( \frac{\sqrt{A_i} - \sqrt{A_{\min}}}{\sqrt{A_{\max}} - \sqrt{A_{\min}}} \right) \cdot 0.9 + 0.1,
\label{eq:gra_calc}
\end{align}
where $A_i$ demonstrates the area of the mask $m_i$ in pixel space, and $A_{\min}$ and $A_{\max}$ correspond to the smallest and largest masks in the current mask hierarchy.

\figPipe{t!}
\begin{wrapfigure}{r}{0.35\linewidth}
\vspace{-8pt}
\captionsetup{font=footnotesize, skip=2pt}
\centering
\includegraphics[width=\linewidth]{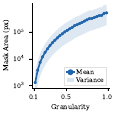}
\vspace{-6pt}
\caption{\textbf{Quantitative analysis on granularity and absolute mask areas.}}
\label{fig:gra-dist-wrap}
\vspace{-12pt}
\end{wrapfigure}
As shown in \cref{fig:sam-gra-dist}(b), our divide-and-conquer pipeline generates a dense hierarchy of masks spanning fine to coarse scales.
We find that most masks have $g_i < 0.4$, indicating an abundance of part-level structures; yet \Ours\ learns stable semantics across all scales, and our granularity is robust to absolute object sizes (Fig.~\ref{fig:gra-1-analysis}). We quantify the relationship between granularity and absolute mask size in Fig.~\ref{fig:gra-dist-wrap} with our unsupervised masks and granularity data in 6,000 images. As granularity increases, the mean mask area grows, showing effective granularity controllability; the variance widens because instance-level masks can differ greatly in area across instances.

\subsection{\ours Architecture}
\label{sec:method-model}
\paragraph{Granularity encoding.}
As shown in Fig.~\ref{fig:pipe}, we introduce a lightweight granularity encoding module that converts a scalar $g$ into a high-dimensional Fourier embedding~\citep{tancik2020fourier}, denoted as $\phi(g) \in \mathbb{R}^{d_{\text{Fourier}}}$.
Then, we project $\phi(g)$ into the decoder feature space using a three-layer MLP:
$
E_g = \text{MLP}(\phi(g)) \in \mathbb{R}^{d_{\text{decoder}}}.
$
This granularity embedding is concatenated with the point prompt features $E_p$:
$
E_{\text{prompt}} = \text{Concat}(E_p, E_g).
$
This design allows the mask decoder to jointly interpret spatial prompts and granularity cues.

\paragraph{Granularity-aware mask decoding.}
We replace SAM’s three fixed-size mask tokens with a single learnable \textit{granularity-aware mask token}.  
This token attends to both prompt embeddings and image features through self- and cross-attention, producing a mask with the target granularity.  
The design is model-agnostic and can be integrated with any promptable segmentation framework.

\paragraph{Parameter efficiency.}
The encoding and token modules introduce less than 0.02\% extra parameters while providing continuous mask granularity control.

\subsection{Distinctions with prior methods.}
\label{sec:method-distinctions}
Prior methods such as GraCo~\citep{zhao2024graco} and GARField~\citep{kim2024garfield} treat granularity as a discrete variable or directly associate it with absolute mask size.
However, discretizing granularity imposes artificial boundaries between adjacent levels, preventing smooth transitions across scales, while absolute-size definitions fail to account for contextual differences, \eg, a ``small'' mask could correspond to an entire object in one image but only a part in another.
In contrast, \Ours\ formulates granularity as a \textit{\textbf{continuous, relative}} measure within the instance–part hierarchy, aligning more closely with human perception, where objects and parts are interpreted in relation to one another rather than by their absolute scales.
This formulation enables \Ours\ to capture fine-grained hierarchical structure and traverse segmentation levels seamlessly.

\figGracoOurs{t!}

\subsection{\Oursplus: Light-Supervised Variant}
\label{sec:method-unsam+}

Although the unsupervised pipeline generates abundant mask–granularity pairs, pseudo labels can be noisy.  
To further stabilize learning, we introduce \textbf{\Oursplus}, a lightly supervised variant that incorporates SA-1B ground-truth masks into the divide stage:
$
\mathcal{M}_{\text{\Oursplus}} = \mathcal{M}_{\text{CutLER}} \cup \mathcal{M}_{\text{SA-1B}}.
$
The combined masks are then processed through the same conquer and granularity-assignment stages as in \Ours.  
This hybrid supervision effectively balances human-curated and self-discovered hierarchies, producing cleaner masks while preserving the scalability and flexibility of unsupervised granularity learning.

\vspace{-8pt}
\subsection{Language-Guided Segmentation with Granularity Control}
\label{sec:method-vlm-gra}

Inspired by previous work in language-guided segmentation, \eg LISA~\citep{lai2024lisa}, Sa2VA~\citep{yuan2025sa2va}, and SAM-3~\citep{carion2025sam}, we integrate an external vision-language model (VLM) to our granularity-controllable segmentation pipeline. Intuitively, we combine \Ours with language since the text semantics also reflect object granularity. The VLM acts as a semantic bridge that predicts the appropriate granularity from a natural point--text prompt. As shown in Fig.~\ref{fig:vlm_pipe}, the user provides an image \({I}\), a click \(p\), and a text description \(t\) of the target, \eg "the cat's left eye." We query a VLM-based predictor \(\phi\) to infer a scalar granularity \(g\), and then prompt UnSAMv2 with \((p,g)\) to produce the final mask:
\begin{equation}
g = \phi(I, p, t), 
\qquad 
\hat{{m}} = f_{\theta}({I}, p, g),
\end{equation}
where \(f_{\theta}\) denotes our granularity-controllable segmentation model.

\figVLMPipe{t!}

To train $\phi$, we leverage the pretrained VLM and our unsupervised mask, granularity pairs $\{(m_i, g_i)\}$ from our divide-and-conquer pipeline in Sec~\ref{sec:method-dico}. For each mask $m_i$, we sample a point $p_i \in m_i$ and associate it with a short phrase $t_i$ describing the region, yielding training tuples $(I, p_i, t_i, g_i)$. We extend the VLM vocabulary with a special \texttt{<gra>} token. During training, the model is prompted to produce a response ending with \texttt{<gra>}, and we attach a lightweight regression head $h_\theta$, a two-layer MLP with a sigmoid output, to the hidden state $\mathbf{z}$ at the \texttt{<gra>} token position. The predicted granularity is $\hat{g} = h_\theta(\mathbf{z})$, and we supervise it with the hierarchy-defined granularity label:
\begin{equation}
\mathcal{L}{\text{gra}} = \left\lVert h_\theta(\mathbf{z}{\texttt{<gra>}}) - g_i \right\rVert_1,
\end{equation}
where $\mathbf{z}{\texttt{<gra>}}$ is the VLM's last-layer hidden representation at the \texttt{<gra>} position, which encodes the full image–point–text context through causal attention. We train the LoRA adapters and $h_\theta$, while the VLM backbone remains frozen. At inference time, the model performs single-shot point-and-text segmentation by autoregressively generating up to the \texttt{<gra>} token, extracting the predicted granularity $\hat{g}$ from the regression head, and then feed the predicted granularity to UnSAMv2 to derive desired segmentation result.

\label{sec:gra-vlm}

\newcommand{\intragroup}{\hspace{1.5pt}}
\newcolumntype{G}{@{}p{6pt}@{}}

\def\tabMainSAMNew#1{%
\begin{table*}[#1]
\tablestyle{0.45pt}{1.0}
\footnotesize
\centering
\resizebox{\textwidth}{!}{%
\begin{tabular}{l
  ccc G ccc G ccc G ccc G ccc G ccc}
Datasets $\rightarrow$ &
\multicolumn{3}{c}{Averaged} & &
\multicolumn{3}{c}{GrabCut} & &
\multicolumn{3}{c}{Berkeley} & &
\multicolumn{3}{c}{SBD} & &
\multicolumn{3}{c}{PascalPart} & &
\multicolumn{3}{c}{PartImageNet} \\
\cline{2-24}
Method &
$\text{NoC}_{80\downarrow}$ & $\text{NoC}_{90\downarrow}$ & 1-IoU & &
$\text{NoC}_{80}$ & $\text{NoC}_{90}$ & 1-IoU & &
$\text{NoC}_{80}$ & $\text{NoC}_{90}$ & 1-IoU & &
$\text{NoC}_{80}$ & $\text{NoC}_{90}$ & 1-IoU & &
$\text{NoC}_{80}$ & $\text{NoC}_{85}$ & 1-IoU & &
$\text{NoC}_{80}$ & $\text{NoC}_{85}$ & 1-IoU \\
\hline
SAM-2~\citep{ravi2024sam} &
4.02 & 5.69 & 58.0 & &
1.48 & 1.54 & 76.8 & &
1.29 & 1.72 & 80.0 & &
2.85 & 6.73 & 64.6 & &
9.37 & 11.80 & 29.0 & &
5.09 & 6.64 & 39.5 \\
\rowcolor{teal!5}
UnSAMv2$^*$ &
\bf 3.41 & \bf 4.75 & \bf 73.1 & &
\bf 1.16 & \bf 1.30 & \bf 91.4 & &
\bf 1.10 & \bf 1.60 & \bf 90.8 & &
\bf 2.75 & \bf 5.97 & \bf 74.8 & &
\bf 7.84 & \bf 9.59 & \bf 51.5 & &
\bf 4.21 & \bf 5.29 & \bf 57.1 \\
\rowcolor{teal!5}
\textit{vs.\ sup.\ SAM-2} &
\spplus{-0.61} & \spplus{-0.94} & \spplus{+15.1} & &
\spplus{-0.32} & \spplus{-0.24} & \spplus{+14.6} & &
\spplus{-0.19} & \spplus{-0.12} & \spplus{+10.8} & &
\spplus{-0.10} & \spplus{-0.76} & \spplus{+10.2} & &
\spplus{-1.53} & \spplus{-2.21} & \spplus{+22.5} & &
\spplus{-0.88} & \spplus{-1.35} & \spplus{+17.6} \\
\hline
\textcolor{OracleTextColor}{UnSAMv2+}$^*$ &
\textcolor{OracleTextColor}{3.30} & \textcolor{OracleTextColor}{4.56} & \textcolor{OracleTextColor}{74.7} & &
\textcolor{OracleTextColor}{1.22} & \textcolor{OracleTextColor}{1.26} & \textcolor{OracleTextColor}{92.6} & &
\textcolor{OracleTextColor}{1.12} & \textcolor{OracleTextColor}{1.37} & \textcolor{OracleTextColor}{91.1} & &
\textcolor{OracleTextColor}{2.55} & \textcolor{OracleTextColor}{5.72} & \textcolor{OracleTextColor}{77.3} & &
\textcolor{OracleTextColor}{7.87} & \textcolor{OracleTextColor}{9.67} & \textcolor{OracleTextColor}{52.0} & &
\textcolor{OracleTextColor}{3.75} & \textcolor{OracleTextColor}{4.77} & \textcolor{OracleTextColor}{60.7} \\
\end{tabular}}
\caption{\textbf{Comparison with SAM-2 on interactive segmentation benchmarks.}
Trained on 6{,}000 images with purely unsupervised pseudo-labels, \ours significantly
outperforms SAM-2. We report promptable segmentation performance in terms of NoC
and 1-IoU across five benchmarks. * Following~\citep{zhao2024graco}, we select the
optimal granularity from 0.1 to 1.0 in steps of 0.1 and report averaged results.}
\label{tab:sam-main}
\end{table*}%
}

\def\tabClickGraCoUnSAM#1{%
\begin{table*}[#1]
\tablestyle{0.45pt}{1.0}
\footnotesize
\centering
\resizebox{\textwidth}{!}{%
\begin{tabular}{l
  ccc G ccc G ccc G ccc G ccc G ccc}
Datasets $\rightarrow$ &
\multicolumn{3}{c}{Averaged} & &
\multicolumn{3}{c}{GrabCut} & &
\multicolumn{3}{c}{Berkeley} & &
\multicolumn{3}{c}{DAVIS} & &
\multicolumn{3}{c}{SA-1B} & &
\multicolumn{3}{c}{PartImageNet} \\
\cline{2-24}
Method &
$\text{NoC}_{80\downarrow}$ & $\text{NoC}_{90\downarrow}$ & 1-IoU & &
$\text{NoC}_{80}$ & $\text{NoC}_{90}$ & 1-IoU & &
$\text{NoC}_{80}$ & $\text{NoC}_{90}$ & 1-IoU & &
$\text{NoC}_{80}$ & $\text{NoC}_{90}$ & 1-IoU & &
$\text{NoC}_{80}$ & $\text{NoC}_{90}$ & 1-IoU & &
$\text{NoC}_{80}$ & $\text{NoC}_{85}$ & 1-IoU \\
\hline
SimpleClick~\citep{liu2023simpleclick} &
3.32 & 4.87 & 60.2 & &
1.32 & 1.48 & 89.6 & &
1.26 & 2.44 & 84.6 & &
2.88 & 5.38 & 75.6 & &
4.80 & 7.27 & 21.9 & &
6.34 & 7.76 & 29.2 \\
GraCo~\citep{zhao2024graco}$^*$ &
2.35 & 3.42 & 74.4 & &
1.24 & 1.32 & 91.1 & &
1.17 & 1.63 & 88.7 & &
2.84 & 5.09 & 74.0 & &
2.41 & 4.01 & 66.2 & &
4.11 & 5.03 & 51.7 \\
SAM-2~\citep{ravi2024sam} &
2.44 & 3.63 & 69.0 & &
1.48 & 1.54 & 76.8 & &
1.29 & 1.72 & 80.0 & &
2.51 & 4.50 & 71.7 & &
\bf 1.82 & 3.76 & 77.1 & &
5.09 & 6.64 & 39.5 \\
\hline
\rowcolor{teal!5}
UnSAMv2$^*$ &
2.28 & 3.40 & 79.3 & &
\bf 1.16 & 1.30 & 91.4 & &
\bf 1.10 & 1.60 & 90.8 & &
2.75 & 4.66 & 84.7 & &
2.18 & 4.16 & 72.4 & &
4.21 & 5.29 & 57.1 \\
\rowcolor{teal!5}
UnSAMv2+$^*$ &
\bf 2.07 & \bf 3.10 & \bf 81.7 & &
1.22 & \bf 1.26 & \bf 92.6 & &
1.12 & \bf 1.37 & \bf 91.1 & &
\bf 2.36 & \bf 4.40 & \bf 85.7 & &
1.87 & \bf 3.69 & \bf 78.4 & &
\bf 3.75 & \bf 4.77 & \bf 60.7 \\
\rowcolor{teal!5}
\textit{vs.\ prev.\ SOTA} &
\spplus{-0.28} & \spplus{-0.32} & \spplus{+7.30} & &
\spplus{-0.02} & \spplus{-0.06} & \spplus{+1.50} & &
\spplus{-0.05} & \spplus{-0.26} & \spplus{+2.40} & &
\spplus{-0.48} & \spplus{-0.69} & \spplus{+11.70} & &
\spplus{-0.54} & \spplus{-0.32} & \spplus{+12.20} & &
\spplus{-0.36} & \spplus{-0.26} & \spplus{+9.00} \\
\end{tabular}}
\caption{\textbf{State-of-the-art performance on interactive segmentation across benchmarks.} Trained with only 6{,}000 images using a combination of supervised and unsupervised labels, \Oursplus demonstrates superior segmentation quality and indicates how effectively unsupervised methods can complement supervised data. Results of~\citep{liu2023simpleclick, zhao2024graco} are reproduced with official code and checkpoints. *: Following~\citep{zhao2024graco}, we select optimal granularity from 0.1 to 1.0 with a step of 0.1 and report average results.}
\label{tab:click-graco-unsamv2-mean}
\end{table*}%
}

\def\tablefontsmall{\fontsize{8}{10}\selectfont}

\def\tabWholeAndScaleSideBySide#1{%
\begin{table}[#1]
\vspace{-4pt}
\centering

\begin{minipage}[t]{0.53\linewidth}\vspace{0pt}
\centering
\begin{minipage}{\linewidth}
\centering
\tablestyle{0.45pt}{1.0}%
\tablefontsmall
\begin{tabular}{lccccc}
Methods & COCO & LVIS & ADE & Entity & SA-1B \\
\hline
SAM~\citep{kirillov2023segment} & 49.6 & 46.1 & 45.8 & 45.9 & 60.8 \\
UnSAM~\citep{wang2024segment}   & 40.5 & 37.7 & 35.7 & 39.6 & 41.9 \\
UnSAM+~\citep{wang2024segment}  & 52.2 & 50.8 & 45.3 & 49.8 & 64.8 \\
\hline
\rowcolor{teal!5} UnSAMv2  & 74.5 & 63.8 & 68.4 & 64.3 & 70.6 \\
\rowcolor{teal!5} UnSAMv2+ & \textbf{79.1} & \textbf{70.3} & \textbf{72.7} & \textbf{70.3} & \textbf{77.9} \\
\rowcolor{teal!5} \textit{vs.\ prev.\ SOTA}
& \spplus{+26.9} & \spplus{+19.5} & \spplus{+27.4} & \spplus{+20.5} & \spplus{+13.1} \\
\end{tabular}
\captionof{table}{\textbf{State-of-the-art performance on whole image segmentation.}
The evaluation metric is $\mathrm{AR}_{1000}$.
For \ours, we aggregate masks at granularity from $0.1$ to $1.0$ in step of $0.1$ and filter out low-confidence masks.}
\label{tab:whole-seg}
\end{minipage}
\end{minipage}%
\hfill
\begin{minipage}[t]{0.42\linewidth}\vspace{0pt}
\centering
\begin{minipage}{\linewidth}
\centering
\tablestyle{0.4pt}{1.0}%
\tablefontsmall
\begin{tabular}{l c c@{\intragroup}c}
Methods & \multicolumn{1}{c}{\# Images~~} &
$\mathrm{NoC}_{80\downarrow}$ & 1-IoU \\
\hline
SAM-2   & -- & 5.09 & 39.5 \\
\hline
UnSAMv2 & 1K & 4.48 & 55.5 \\
UnSAMv2 & 3K & 4.27 & 55.7 \\
\rowcolor{teal!5}%
UnSAMv2 & 6K & 4.21 & 57.1 \\
\end{tabular}
\captionof{table}{\textbf{Ablations for training data size.}
\ours starts to demonstrate a decent understanding of granularity when trained with only 1{,}000 images with unsupervised pseudo-labels.
We report metrics on PartImageNet.}
\label{tab:data-scale}
\end{minipage}
\end{minipage}
\vspace{-6pt}
\end{table}%
}

\providecommand{\intragroup}{\hspace{6pt}}

\def\figVideoDemoImg{%
  \centering
  \includegraphics[width=\linewidth]{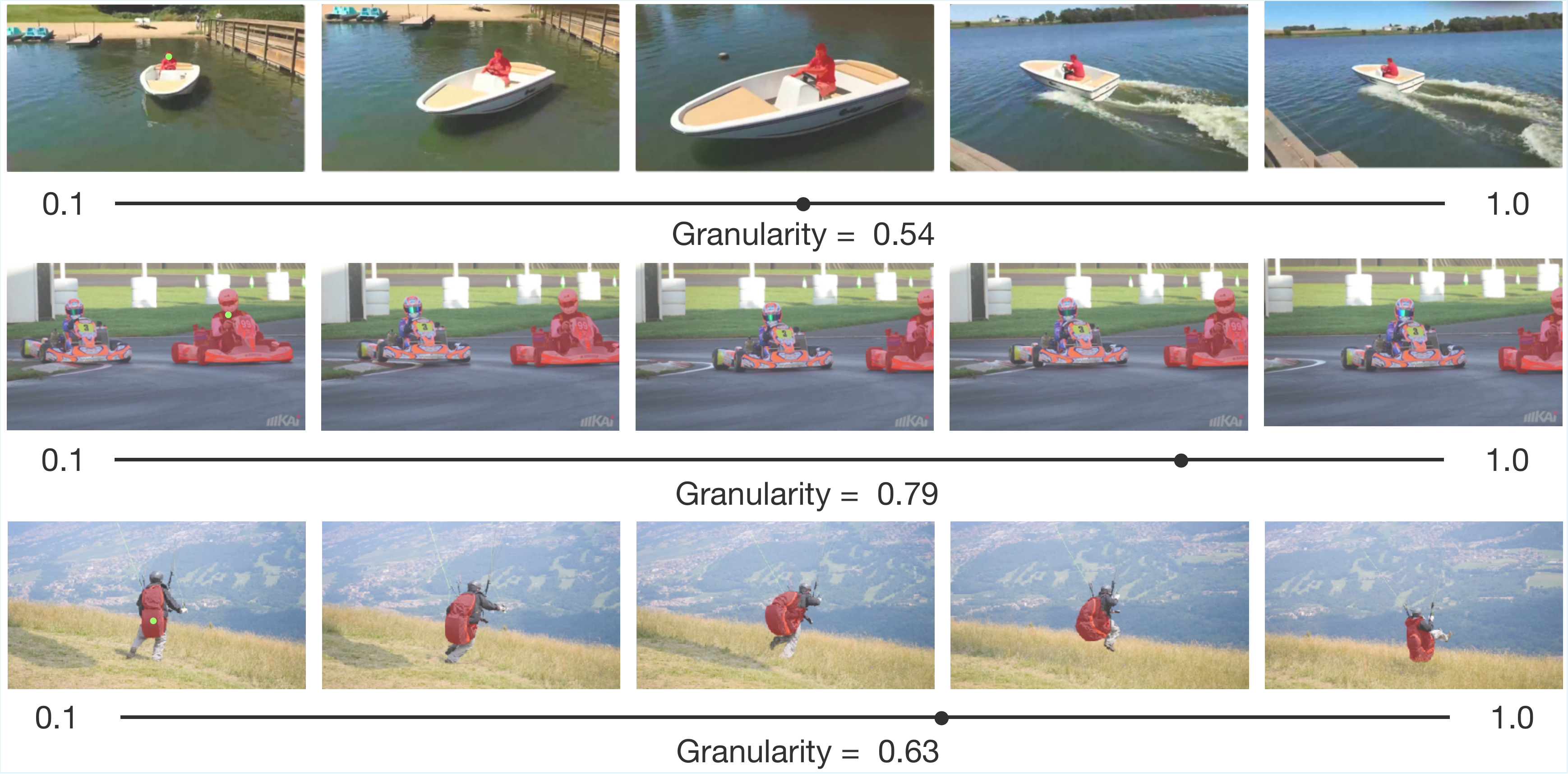}
  \caption{\textbf{Granularity generalizes to video.} We prompt \ours with a point and granularity value on the first frame, then propagate the mask to later frames. Even though trained only on images, \ours maintains consistent masks over time, showing strong temporal coherence and transferability.}
  \label{fig:video_demo}
}

\def\tabVLMCompareBody{%
\tablestyle{0.4pt}{1.0}%
\tablefontsmall
\centering
{\setlength{\tabcolsep}{1pt}%
\begin{tabular}{l@{\hspace{2pt}}l@{\hspace{4pt}}r}
Method & Base VLM & 1-IoU \\
\hline
LISA   & LLaVA~1.5             & 26.31 \\
Sa2VA  & Qwen2.5-VL-7B         & 40.41 \\
\rev{Sa2VA}  & \rev{Qwen3-VL-4B} & \rev{41.25} \\
\hline
\rowcolor{teal!5} UnSAMv2 & LLaVA~1.5           & 43.73 \\
\rowcolor{teal!5} UnSAMv2 & Qwen2.5-VL-7B       & 45.07 \\
\rowcolor{teal!5} \rev{UnSAMv2} & \rev{Qwen3-VL-4B} & \rev{44.52} \\
\rowcolor{teal!5} UnSAMv2 & tuned Qwen2.5-VL-7B & \textbf{47.29} \\
\end{tabular}}%
\captionof{table}{\textbf{Evaluation results on language-guided segmentation.}
We use the PartImageNet~\citep{he2022partimagenet} test set, querying LISA~\citep{lai2024lisa}, Sa2VA~\citep{yuan2025sa2va}, and \Ours with the ground-truth text label and object center point.
Without any finetuning, the external VLMs interpret the granularity scale effectively, suggesting a strong alignment between our granularity and semantics.}%
\label{tab:vlm-compare}%
}

\def\tabfigVideoVLM#1{%
\begin{figure}[#1]
  \centering
  \begin{minipage}[t]{0.5\textwidth}\vspace{0pt}
    \centering
    \figVideoDemoImg
  \end{minipage}
  \hfill
  \begin{minipage}[t]{0.46\textwidth}\vspace{0pt}
    \centering
    \tabVLMCompareBody
  \end{minipage}
\end{figure}%
}

\def\figGraTokenMini{%
  \captionsetup{font=small}
  \centering
  \includegraphics[width=\linewidth]{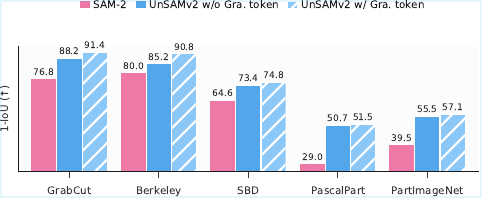}\vspace{-1pt}
  \captionof{figure}{\textbf{Ablation of granularity-aware mask token.}
  Directly finetuning original SAM-2 mask tokens leads to limited gains, suggesting they already encode strong priors.
  Adding our granularity-aware token enables efficient learning of granularity.}
  \label{fig:gra-token-abla}
}

\def\tabSAGTMini{%
  \centering
  \begingroup
  \footnotesize
  \setlength{\tabcolsep}{4pt}
  \renewcommand{\arraystretch}{0.95}

  \begin{adjustbox}{max width=\linewidth}
  \begin{tabular}{l c c cc}
  \multirow{2}{*}{Methods} &
  \multicolumn{1}{c}{Sup.} &
  \multicolumn{1}{c}{Unsup.} &
  \multicolumn{2}{c}{1-IoU} \\
  \cline{4-5}
  & \multicolumn{1}{c}{Data} & \multicolumn{1}{c}{Data} &
  PascalPart & PtIn \\
  \hline
  SAM-2    & --     & --     & 29.0 & 39.5 \\
  \hline
  UnSAMv2  & \cmark & \xmark & 42.5 & 54.3 \\
  \rowcolor{teal!5}%
  UnSAMv2  & \xmark & \cmark & 51.5 & 57.1 \\
  \rowcolor{teal!5}%
  UnSAMv2+ & \cmark & \cmark & 52.0 & 60.7 \\
  \end{tabular}
  \end{adjustbox}

  \endgroup

  \captionof{table}{\textbf{Ablation for supervised granularity training.}
  Training with only SA-1B supervised labels underperforms using data from our unsupervised pipeline, highlighting the importance of pseudo-labels for learning granularity.}
  \label{tab:sa1b-gt}
}

\def\figtabGraTokenSAGT#1{%
  \begin{figure}[#1]
    \captionsetup{font=small, skip=2pt}
    \centering

    \begin{minipage}[t]{0.60\linewidth}\vspace{0pt}
      \captionsetup{width=\linewidth}
      \figGraTokenMini
    \end{minipage}
    \hfill
    \begin{minipage}[t]{0.38\linewidth}\vspace{0pt}
      \captionsetup{width=\linewidth}
      \tabSAGTMini
    \end{minipage}

  \end{figure}%
}
\vspace{-8pt}
\section{Experiments}

\subsection{Model Training Settings}
\textbf{Pseudo mask-granularity data generation.} For the divide stage of our unsupervised data pipeline, we set $\tau_{\text{conf}}$ to 0.3 and $\tau_{\text{overlap}}$ to 0.8. In the conquer stage, we leverage DINOv3~\citep{simeoni2025dinov3} vision transformer to extract feature embedding and set the iterative merging thresholds as $\theta = [0.9, 0.8, 0.7, 0.6, 0.5]$. We run the unsupervised pipeline to generate mask-granularity pairs on 6,000 images from SA-1B. 
\textbf{\ours training.} We adopt SAM-2-small as the base model, and with only 8 A-100 GPU hours, we finetune the base model for 5 epochs with 6,000 unlabeled images. During training, we freeze the vision encoder and train the granularity encoding module, granularity mask token, and the two-way transformer block in the mask decoder with LoRA~\citep{hu2022lora}. \textbf{VLM training.} We adopt Qwen2.5-VL-7B as our base model and use image-text labels from PascalPart and PartImageNet as our training data. We train the base model for 2 epochs with LoRA. We provide more experimental details in the appendix.

\subsection{Evaluation Datasets and Metrics}
\noindent \textbf{Interactive Image Segmentation.} Following SimpleClick~\citep{liu2023simpleclick} and GraCo~\citep{zhao2024graco}, we evaluate \ours's capability to segment objects at various granularity levels with the number of clicks (NoC) required to reach a certain Intersection over Union (IoU) threshold to assess the model's efficiency in segmenting the target object. Specifically, we choose $\text{NoC}_{80} \text{ and } \text{NoC}_{90}$ which record the average number of clicks needed to achieve 80\% and 90\% IoU between predicted mask and ground-truth mask. In addition, we measure the IoU between the predicted mask and the ground-truth mask with just one click, denoted as 1-IoU in the table. For object levels, we conduct evaluations on 5 commonly used datasets~\citep{rother2004grabcut, mcguinness2010comparative, perazzi2016benchmark, hariharan2011semantic, kirillov2023segment}, 
and for part level, we evaluate models on~\citep{chen2014detect, imagenet}.
Note that for the fairness of comparison, we only compare datasets across models that are not in-distribution with their training data, e.g., SAM-2~\citep{ravi2024sam} is trained on SA-1B, SimpleClick~\citep{liu2023simpleclick} and GraCo~\citep{zhao2024graco} are trained with SBD and PascalPart. We provide descriptions of evaluation datasets in the appendix.
\par\noindent \textbf{Whole-Image Segmentation.}
We evaluate our model's performance in identifying all possible masks for given images in a zero-shot setting across 5 datasets that span a broad range of granularity~\citep{lin2014microsoft, gupta2019lvis, zhou2019semantic, qi2022open, kirillov2023segment}.
Importantly, each benchmark annotates only specific semantic classes and typically emphasizes certain levels of the part–instance hierarchy, whereas our method predicts masks at all levels and for any class. Consequently, COCO Average Precision (AP) does not faithfully capture open-world performance. Following prior work~\citep{wang2023cut,cao2024sohes,wang2024segment}, we report Average Recall ($\mathrm{AR}_{1000}$) for comparison across methods.

\par\noindent \textbf{Language-Guided Segmentation.} To evaluate our text-guided segmentation pipeline in~\cref{sec:gra-vlm}, we test \Ours augmented with an external VLM on PartImageNet~\citep{he2022partimagenet}. For each test image, we take the center point of the ground-truth mask and its text label and query the VLM to predict a granularity value $g \in [0.1, 1.0]$. We then prompt \Ours with the same center point and the predicted granularity to produce the segmentation mask and report 1-IoU. For fair comparison, we provide the same center-point and text-label inputs to the baselines LISA~\citep{lai2024lisa} and Sa2VA~\citep{yuan2025sa2va}.
\vspace{-8pt}
\subsection{Experimental Results}
\tabMainSAMNew{t!}
\tabClickGraCoUnSAM{t!}

\textbf{Interactive Segmentation.} Remarkably, \ours surpasses SAM-2 across {all} datasets in zero-shot evaluation settings as summarized in Table~\ref{tab:sam-main}. Finetuned with only 6,000 images with unsupervised pseudo-labels, \ours demonstrates superior performance in segmenting objects at various granularity levels. It indicates that our model architecture and unsupervised data pipeline effectively guide the base model to understand the meaning of granularity scaler. On average, \ours surpasses SAM-2 by 15.2\% in $\text{NoC}_{80}$, 16.5\% in $\text{NoC}_{90}$, and 26.0\% in 1-IoU. With \ours, user could segment their desired objects accurately at any granularity level with fewer prompt points, which greatly enhances the flexibility of object segmentation and prepares for downstream tasks. In addition, as shown in Tab.~\ref{tab:click-graco-unsamv2-mean}, \Oursplus, trained with a combination of supervised and unsupervised labels on 6,000 images achieves state-of-the-art performance on all metrics across multiple evaluation datasets.

Qualitative comparisons with previous SOTA~\citep{zhao2024graco} are shown in Fig.~\ref{fig:graco-ours}. For fair comparison with the three mask tokens design in SAM-2, we additionally report results using three granularity scales 0.2, 0.5, and 1.0 in Table~\ref{tab:sam2_unsamv2_gra_ablation}.

\tabWholeAndScaleSideBySide{t}
\parImperfectGWrapBlock

\begin{wraptable}{r}{0.33\linewidth}
\vspace{-8pt}
\captionsetup{font=footnotesize, skip=2pt}
\centering
\scriptsize
\setlength{\tabcolsep}{1.8pt}
\renewcommand{\arraystretch}{0.90}

\begin{tabular}{@{}l@{\hspace{2pt}}c@{\hspace{3pt}}c@{}}
\hline
Method & NoC$_{80}\downarrow$ & 1-IoU \\
\hline
SAM-2 (3 tok.)  & 4.02 & 58.0 \\
\rowcolor{teal!5}
UnSAMv2 (3 gra.)  & \underline{3.76} & \underline{66.7} \\
\rowcolor{teal!5}
UnSAMv2 (10 gra.) & \bf 3.41 & \bf 73.1 \\
\hline
\end{tabular}

\caption{\textbf{Evaluation with three granularity scales.}}
\label{tab:sam2_unsamv2_gra_ablation}
\vspace{-8pt}
\end{wraptable}

\noindent \textbf{Whole-Image Segmentation.}
Apart from SOTA results in interactive segmentation, \ours achieves superior performance on whole-image segmentation across datasets with instances of abundant granularity levels, outperforming SAM~\citep{kirillov2023segment} by 37.7\% and UnSAM~\citep{wang2024segment} by 29.8\% in $\text{AR}_{1000}$. The results are shown in Table~\ref{tab:whole-seg}. With a granularity scalar as input, \ours can surface instances over a wide range of detail. As shown in Fig.~\ref{fig:ours-whole-seg}, users simply set the desired granularity to obtain all candidate masks in images at that level, creating new possibilities on how to integrate segmentation models into vision task pipelines.
\figOursWholeSeg{t!}

\tabVideoQuantWrap
\noindent \textbf{Video Segmentation Results.}
Despite being trained solely on images, \ours demonstrates promising capabilities on interactive video segmentation. Although we keep SAM-2’s memory module frozen during training, \ours still achieves competitive results on video data, demonstrating that the granularity embeddings and mask token propagate across frames effectively. This further shows that our self-supervised pipeline enables granularity to be assimilated into pretrained model's reasoning flow. We show qualitative video results in Fig.~\ref{fig:video_demo}.
\rev{Quantitatively, Table~\ref{tab:video-quant} evaluates on
DAVIS~\citep{perazzi2016benchmark}. We feed granularity to select the first-frame mask and SAM-2's frozen memory module propagates it to later frames. \ours improves J\&F, anchored J\&F, and temporal-consistency error over SAM-2, showing that granularity-selected first-frame masks transfer to video, while temporal supervision of granularity scales remains a direction for future work.}

\noindent \textbf{Language-Guided Segmentation Results.} By integrating an external VLM into our granularity-controllable segmentation pipeline, we leverage the strong correlation between granularity and text semantics to further unlease the potential of \Ours. The VLM can reason on what a reasonable granularity scale is given a short text description of the desired object or part and works as a bridge between users' text prompts and granularity scales. By combining \Ours with an external VLM, we achieve promising results compared with widely used language segmentation models~\citep{lai2024lisa, yuan2025sa2va} as demonstrated in Table~\ref{tab:vlm-compare}.

\tabfigVideoVLM{t}

\section{Ablations and Analysis}

\noindent\textbf{Efficiency of \ours training.}
In Table~\ref{tab:data-scale}, we ablate the training data size used for \ours. We find that granularity training is highly sample-efficient: \ours already shows a solid grasp of granularity with only 1{,}000 images. We attribute this efficiency to two factors. First, we derive the granularity scale hierarchically, mirroring how humans perceive scale—by relative size within a hierarchy rather than absolute size. Second, during training we update only $0.1\%$ of the parameters and keep the rest frozen, which preserves pretrained model's segmentation ability. Overall, our procedure effectively teaches pretrained model the concept of mask granularity: with just a few example images, \ours learns a latent representation of granularity rather than merely memorizing instances.

\noindent\textbf{Granularity-aware mask token.} We observe that training \ours directly with the original mask tokens and token decoding module in SAM-2 leads to unsatisfying performance (Fig.~\ref{fig:gra-token-abla}). This phenomenon has been noticed in HQ-SAM~\citep{ke2023segment}. It indicates SAM-2's original mask tokens have already shown strong-prior knowledge on what constitutes as an object. Teaching these pretrained tokens to understand the meaning of granularity is difficult. Thus, we introduce a new mask token and its MLP to do mask decoding that is only trained with mask data accompanied with granularity scales. By conducting self-attention with the granularity embedding encoded by Fourier module and cross-attention with image embeddings, the newly introduced token learn the representation of masks and their corresponding granularity simultaneously. The newly introduced token is a crucial component of \ours and a key component to achieve the goal -- segment anything at any granularity.
\figtabGraTokenSAGT{t}

\noindent\textbf{Unsupervised pseudo-labels} are pivotal for granularity training, as shown in Table~\ref{tab:sa1b-gt}. We observe that when trained solely on human-labeled SA-1B annotations, \ours performs unsatisfactorily compared with training that also incorporates pseudo-labels from our divide-and-conquer pipeline. This points to an inherent issue with supervised datasets: they are heavily biased by human labelers’ notions of objects. By contrast, our unsupervised approach focuses on intrinsic relationships among patches, enabling coverage of both instances and parts hierarchically.

\tabRepPseudoSubtables{t}
\parSemanticsWithTable

\rev{\noindent\textbf{Granularity representation and pseudo-label design.}
Table~\ref{tab:gra-rep-ablation} ablates the continuous Fourier-encoded
granularity representation against discrete bins and a plain MLP projection. We 
keep architecture, training data, and hyperparameters fixed.
Table~\ref{tab:pseudo-label-ablation} ablates the pseudo-label design, which shows that hierarchy-based granularity assignment, square-root scaling, and the conquer stage are all necessary.}
\vspace{-8pt}

\section{Conclusion}
\vspace{-3pt}
We presented \ours, a self-supervised framework that equips pretrained segmentation model to segment anything at any granularity.
By deriving continuous granularity scales from unlabeled data, \ours learns to traverse part–whole hierarchies and control segmentation with a single scalar.
Trained on only 6K unlabeled images, it achieves state-of-the-art results across interactive, whole-image, and video segmentation. We further combine our granularity pipeline in language-guided segmentation, demonstrating the wide usage of our granularity pipeline.
Our results highlight that self-supervised learning can unlock latent hierarchical structure in vision foundation models, transforming segmentation from discrete prediction into continuous, controllable reasoning.

\newpage

\hbadness=1500
\bibliographystyle{tmlr}
\bibliography{main}

\appendix
\def\figIntDemoApp#1{
    \begin{figure*}[#1]
      \centering
      \includegraphics[width=0.99\linewidth]{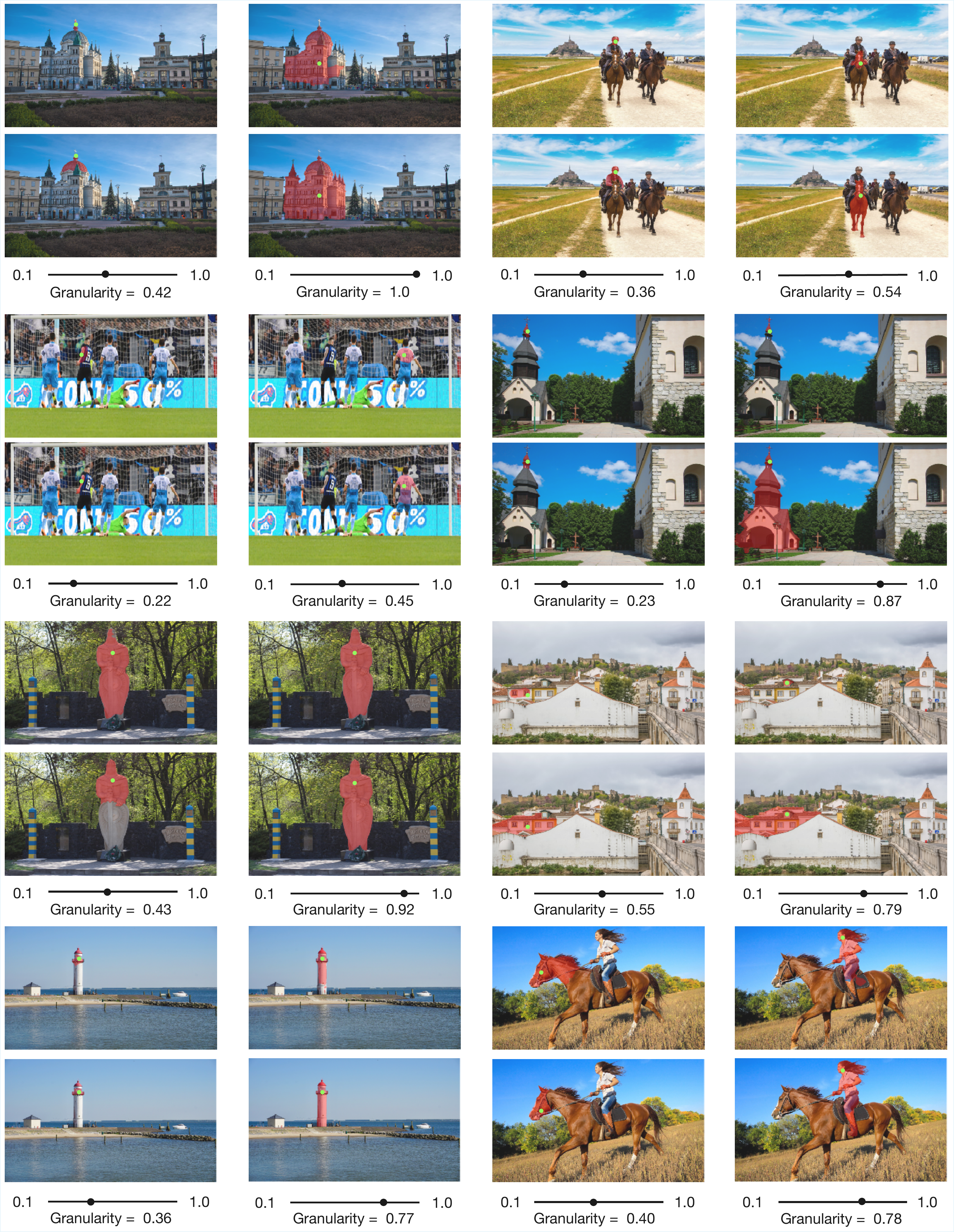} \vspace{-6pt}
    \caption{Interactive segmentation results on SA-1B. The top row is previous SOTA GraCo~\citep{zhao2024graco} and the bottom row is \ours.}
    \label{fig:app-int}
    \end{figure*}
}

\def\figWholeDemoApp#1{
    \begin{figure*}[#1]
      \centering
      \includegraphics[width=0.99\linewidth]{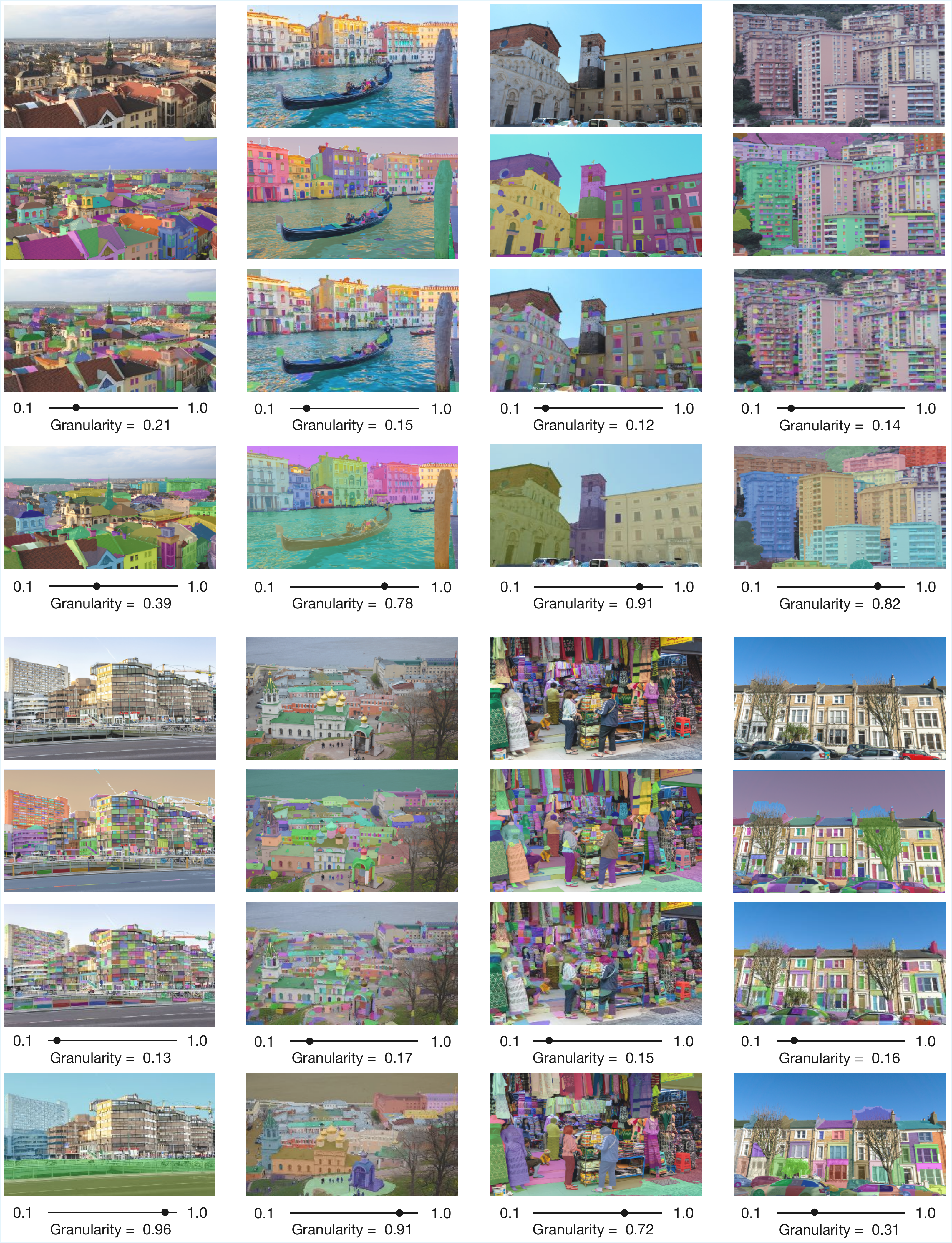} \vspace{-6pt}
    \caption{Whole image segmentation on SA-1B. From top to bottom are raw images, segmentation by SAM-2 and \ours.}
    \label{fig:app-whole}
    \end{figure*}
}

\def\figVideoDemoApp#1{
    \begin{figure*}[#1]
      \centering
      \includegraphics[width=0.99\linewidth]{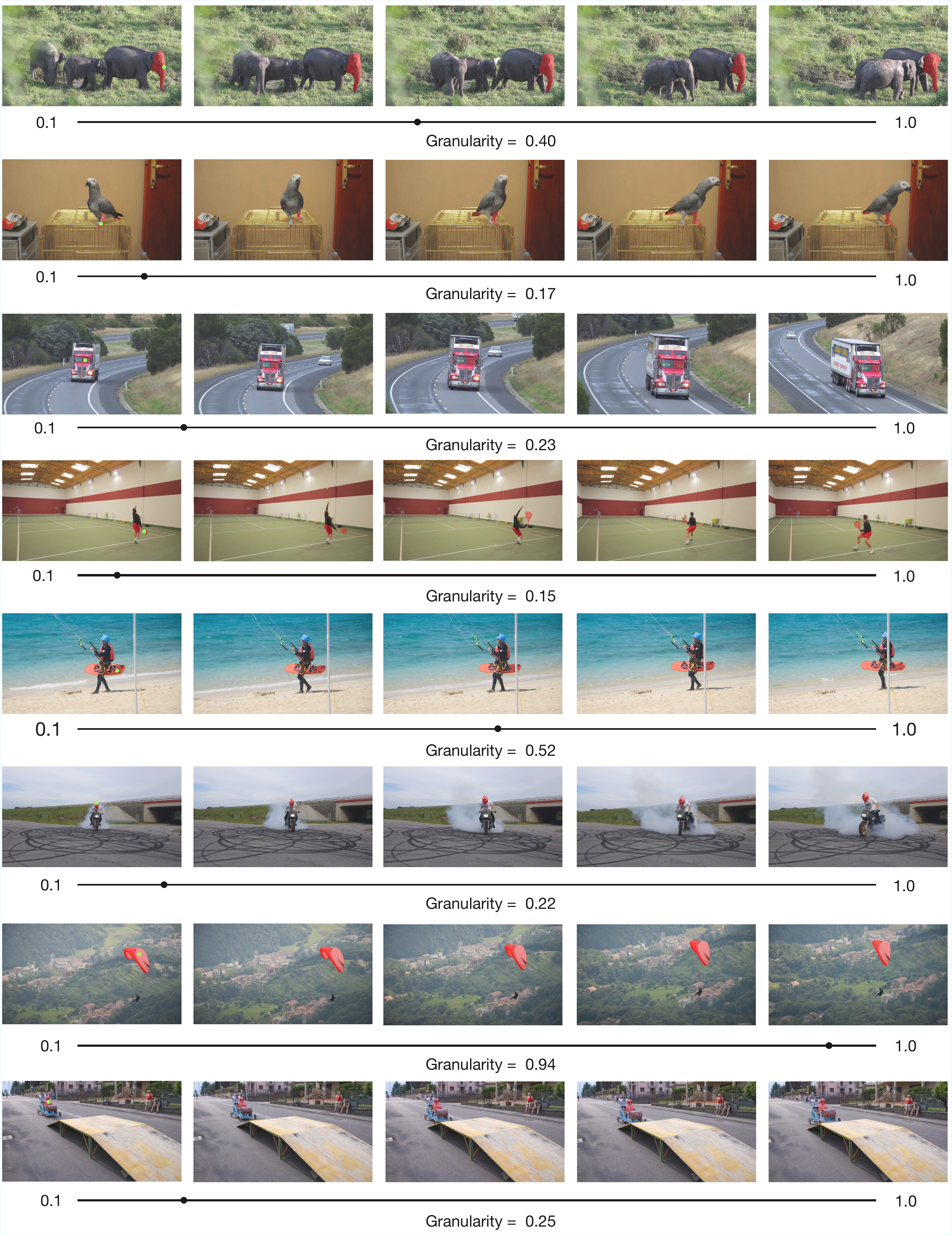}\vspace{-3pt}
    \caption{\ours's interactive video segmentation results on YoutubeVIS dataset at various granularity levels.}
    \label{fig:app-video}
    \end{figure*}
}

\clearpage
\renewcommand{\thefigure}{A\arabic{figure}}
\setcounter{figure}{0}
\renewcommand{\thetable}{A\arabic{table}}
\setcounter{table}{0}
\renewcommand{\thesection}{A\arabic{section}}
\setcounter{section}{0}

\section{Evaluation Datasets}
\label{app:eval-data}
\paragraph{Interactive Segmentation.} Specifically, we adopt the following 7 datasets as benchmarks.

\begin{itemize}
    \item \textbf{GrabCut}~\citep{rother2004grabcut} consists of 50 images, each containing a single instance. 
    \item \textbf{Berkeley}~\citep{mcguinness2010comparative} includes 96 images with 100 instances, some of which are more challenging for segmentation.
    \item \textbf{DAVIS}~\citep{perazzi2016benchmark} contains 50 high-quality videos and we use 345 frames for evaluation.
    \item \textbf{SA-1B}~\citep{kirillov2023segment} contains 11 million high-resolution images (average size 3300×4950 pixels) and 1.1 billion high-quality segmentation masks. For evaluation, we randomly choose 1,000 images that are not included in \ours training data as our evaluation set.
    \item \textbf{SBD}~\citep{hariharan2011semantic} includes 8,498 training images (with 20,172 instances) and 2,857 validation images (with 6,671 instances). 
    \item \textbf{PascalPart}~\citep{chen2014detect} provides part-level annotations for 20 classes from PascalVOC, totaling 193 part categories. We evaluate models on the whole validation set. 
    \item \textbf{PartImageNet}~\citep{imagenet} organizes 158 ImageNet classes into 11 super-categories and defines 40 distinct part categories. We utilize its validation set to evaluate model's performance on segmenting part-level objects, which consists of 1,206 images and 4,553 annotated parts.
\end{itemize}

\paragraph{Whole-Image Segmentation.} We adopt the following 5 datasets to evaluate \ours's capability on discovering all instances in images.
\begin{itemize}
    \item \textbf{COCO} (Common Objects in Context)~\citep{lin2014microsoft} is a widely utilized object detection and segmentation dataset. It consists of 115,000 labeled training images and 5,000 labeled validation images. We evaluate our model on COCO \texttt{Val2017} with 5000 validation images in a zero-shot manner. We use averaged recall ($\text{AR}_{1000}$) as the metrics for the whole-image segmentation task.
    
    \item \textbf{SA-1B}~\citep{kirillov2023segment} consists of 11 million high-resolution images and 1.1 billion segmentation masks. Following interactive segmentation, we randomly choose 1,000 images that are not included in \ours training procedure as evaluation set.

    \item \textbf{LVIS} (Large Vocabulary Instance Segmentation)~\citep{gupta2019lvis} has 164,000 images with over 1,200 categories and 2 million high-quality instance-level segmentation masks. It covers a large number of object categories. We evaluate \ours using its 5000 validation images.

    \item \textbf{EntitySeg}~\citep{qi2022open} is an open-world, class-agnostic dataset with 33{,}277 images, averaging 18.1 annotated entities per image. We conduct zero-shot evaluation on 1{,}314 low-resolution images in the validation set.
    
    \item \textbf{ADE20K}~\citep{zhou2019semantic} contains 25{,}574 training and 2{,}000 testing images covering 365 scenes, emphasizing semantic-level segmentation. It provides labels for 150 semantic categories and 707{,}868 objects drawn from 3{,}688 categories. We evaluate zero-shot whole-image segmentation performance on the 2{,}000-image test split.
\end{itemize}

\paragraph{Language-Guided Segmentation.} We use the test of PartImageNet~\citep{he2022partimagenet} as our evaluation data. 
For each sample, the VLM receives the image and a text description of the target part and predicts a granularity score. A center click point is simulated on the ground-truth mask, and the predicted granularity along with the point prompt are passed to \ours. We report mean IoU at 1-click.

\section{Training Details.}
\label{app:train-detail}
\paragraph{Pseudo mask-granularity data generation.} For the divide stage of our unsupervised data pipeline, we set $\tau_{\text{conf}}$ to 0.3 and $\tau_{\text{overlap}}$ to 0.8. In the conquer stage, we leverage DINOv3~\citep{simeoni2025dinov3} ViT-B/16 backbone to extract feature embedding from the last layer of vision transformer and merge adjacent patches together based on predefined cosine similarity thresholds $\theta = [0.9, 0.8, 0.7, 0.6, 0.5]$. We run the pipeline to generate mask-granularity pairs on 6,000 images from SA-1B in a fully unsupervised manner. On average, we have 112 pseudo-labels on each image. Note that unlike UnSAM~\citep{wang2024segment}, \ours is designed to learn instance–part relationships rather than only instance representations, so our granularity-aware divide-and-conquer pipeline intentionally produces fewer pseudo-labels than UnSAM, which produces 448 pseudo-labels per image.

\paragraph{\ours Training.} We finetune SAM-2-small model for 5 epochs with pseudo-labels on 6,000 images. During finetuning, we freeze the heavy-weight Hiera image encoder and only train the granularity encoding module, granularity mask token, and the two-way transformer block in the mask decoder. For the granularity encoder, we first adopt Fourier transformation to granularity scalar with dimension $d_\text{Fourier} = 128$ and followed by a 3-layer MLP. Following SAM-2~\citep{ravi2024sam}, \ours adopts a combination of focal loss and dice loss with a ratio of 20:1. The batch size is set to 4, with the learning rate initialized at 1e-4. We apply LoRA technology to all projection layers of the transformer, setting the LoRA rank to 8. All experiments are conducted on either 2 A-100 or 4 RTX 3090 GPUs. 

\paragraph{VLM training.} We construct the granularity training dataset from PascalPart and PartImageNet. Each sample consists of an image, a natural language description of the target part (e.g., "head of the person"), and a ground-truth granularity score in [0.1, 1.0] calculated with our divide-and-conquer pipeline. We adopt Qwen2.5-VL-7B-Instruct as the VLM backbone. The model is trained with LoRA with rank set to 16. A special \texttt{<gra>} token is appended to each response; its last-layer hidden state is fed through a lightweight regression head to produce the granularity prediction. The model is optimized with AdamW using a cosine learning rate schedule. All training is conducted on 2 A-100 GPUs.

\secRebuttalAppendix

\section{More Visualizations}
We present more \ours's qualitative results on interactive image segmentation in Fig.~\ref{fig:app-int}, whole image segmentation in Fig.~\ref{fig:app-whole}, and video segmentation in Fig.~\ref{fig:app-video}.
\clearpage
\figIntDemoApp{p}
\figWholeDemoApp{p}
\figVideoDemoApp{p}

\end{document}